\documentclass[10pt,journal,compsoc]{IEEEtran}
\ifCLASSOPTIONcompsoc
  \usepackage[nocompress]{cite}
\else
  \usepackage{cite}
\fi
\usepackage{graphicx}
\graphicspath{{./figures/}} %
\usepackage{amsmath}
\usepackage{amsfonts}
\usepackage{amssymb}

\usepackage[ruled,vlined]{algorithm2e}
\usepackage{array}
\ifCLASSOPTIONcompsoc
 \usepackage[caption=false,font=footnotesize,labelfont=sf,textfont=sf]{subfig}
\else
 \usepackage[caption=false,font=footnotesize]{subfig}
\fi
\usepackage{url}

\DeclareMathOperator*{\argmax}{\arg\max}
\DeclareMathOperator*{\argmin}{\arg\min}

\usepackage{mathtools}
\DeclarePairedDelimiterX{\infdivx}[2]{(}{)}{%
  #1\;\delimsize\|\;#2%
}

\newcommand{\kld}{D_{KL}\infdivx}

\usepackage{booktabs}
\usepackage{multirow}
\usepackage{arydshln}

\begin{document}
\title{CALDA: Improving Multi-Source Time Series Domain Adaptation with Contrastive Adversarial Learning}

\author{Garrett~Wilson,
        Janardhan~Rao~Doppa,~\IEEEmembership{Senior Member,~IEEE}
        and~Diane~J.~Cook,~\IEEEmembership{Fellow,~IEEE}%
\IEEEcompsocitemizethanks{\IEEEcompsocthanksitem The authors are with the School of Electrical Engineering and Computer Science, Washington State University, Pullman, WA, 99164. E-mail: \{garrett.wilson,jana.doppa,djcook\}@wsu.edu}%
}%

\IEEEtitleabstractindextext{%
\begin{abstract}
Unsupervised domain adaptation (UDA) provides a strategy for improving machine learning performance in data-rich (target) domains where ground truth labels are inaccessible but can be found in related (source) domains. In cases where meta-domain information such as label distributions is available, weak supervision can further boost performance. We propose a novel framework, CALDA, to tackle these two problems. CALDA synergistically combines the principles of contrastive learning and adversarial learning to robustly support multi-source UDA (MS-UDA) for time series data. Similar to prior methods, CALDA utilizes adversarial learning to align source and target feature representations. Unlike prior approaches, CALDA additionally leverages cross-source label information across domains. CALDA pulls examples with the same label close to each other, while pushing apart examples with different labels, reshaping the space through contrastive learning. Unlike prior contrastive adaptation methods, CALDA requires neither data augmentation nor pseudo labeling, which may be more challenging for time series. We empirically validate our proposed approach. Based on results from human activity recognition, electromyography, and synthetic datasets, we find utilizing cross-source information improves performance over prior time series and contrastive methods. Weak supervision further improves performance, even in the presence of noise, allowing CALDA to offer generalizable strategies for MS-UDA.
\end{abstract}

\begin{IEEEkeywords}
Transfer Learning, Domain Adaptation, Time Series, Weak Supervision, Adversarial Training, Contrastive Learning.
\end{IEEEkeywords}}

\maketitle

\IEEEdisplaynontitleabstractindextext

\IEEEpeerreviewmaketitle

\IEEEraisesectionheading{\section{Introduction}\label{sec:introduction}}

\IEEEPARstart{U}{nsupervised} domain adaptation can leverage labeled data from past (source) machine learning tasks when only unlabeled data are available for a new related (target) task \cite{wilson2020survey}. As an example, when learning a model to recognize a person's activities from time-series sensor data, standard learning algorithms will face an obstacle when person $A$ does not provide ground-truth activity labels for their data. If persons $B$ through $G$ are willing to provide these labels, then Multi-Source Unsupervised Domain Adaptation (MS-UDA) can create a model for the target person based on labeled data from the source persons.
When performing adaptation, MS-UDA must bridge a domain gap. In our example, such a gap exists because of human variability in how activities are performed. Meta-domain information may exist for person $A$ that is easier to collect and can improve the situation through weak supervision \cite{wilson2020codats}, such as self-reported frequencies for each activity (e.g., ``I sleep 8 hours each night'').

In this article, we develop a framework that can construct a model for time-series MS-UDA. Our proposed approach leverages labeled data from one or more source domains, unlabeled data from a target domain, and optional target class distribution. Very few domain adaptation methods handle time series \cite{purushotham2017variational,wilson2020codats,wilson2020survey} and even fewer facilitate multiple source domains or weak supervision \cite{wilson2020codats}. We postulate adapting multiple time-series domains is particularly critical because many time-series problems involve multiple domains (in our example, multiple people) \cite{anguita2013public,zhao2017icml}. Furthermore, we posit that existing approaches to adaptation do not make effective use of meta-domain information about the target, yet additional gains may stem from leveraging this information via weak supervision. We propose a novel framework for time series MS-UDA. This framework, called CALDA (\textbf{C}ontrastive \textbf{A}dversarial \textbf{L}earning for Multi-Source Time Series \textbf{D}omain \textbf{A}daptation), improves unsupervised domain adaptation through adversarial training, contrastive learning, and weak supervision without relying heavily on data augmentation or pseudo labeling like prior image-based methods.

First, CALDA guides adaptation through multi-source domain-adversarial training \cite{ganin2016jmlr,wilson2020codats}. CALDA trains a multi-class domain classifier to correctly predict the original domain for an example's feature representation while simultaneously training a feature extractor to incorrectly predict the example's domain. Through this two-player game, the feature extractor produces domain-invariant features. A task classifier trained on this domain-invariant representation can potentially transfer its model to a new domain because the target features match those seen during training, thus bridging the domain gap. CALDA utilizes adversarial training to align the feature-level distributions between domains and utilizes contrastive learning to leverage cross-source label information for improving accuracy on the target domain.

Second, CALDA enhances MS-UDA through contrastive learning across source domains. Contrastive learning moves the representations of similar examples close together and
dissimilar examples far apart. While this technique yielded performance gains for self-supervised \cite{chen2020simple,he2020moco} and traditional supervised learning \cite{khosla2020supervisedcontrastive}, the method is unexplored for multi-source time series domain adaptation. We propose to introduce the contrastive learning principle within CALDA. In this context, we will investigate three design decisions. First, we will analyze methods to select pairs of examples from source domains. Second, we will determine whether it is beneficial to pseudo-label and include target data as a contrastive learning domain despite the likelihood of incorrect pseudo-labels due to the large domain shifts in time series data. Third, we will assess whether to randomly select contrastive learning examples or utilize the varying complexities of different domains to select the most challenging (i.e., hard) examples.

\begin{figure}
\centering
\includegraphics[width=0.7\linewidth]{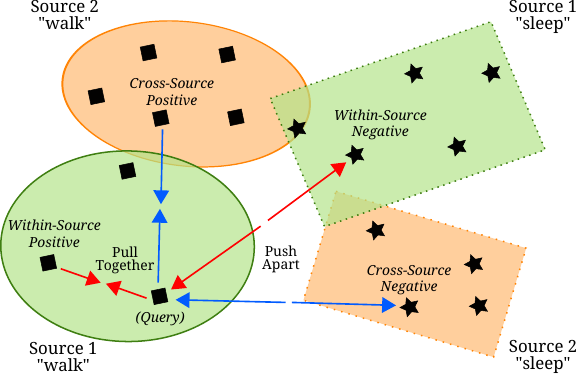}
\caption[Cross-source vs. within-source contrastive learning]{In the
space $z = Z(F(x))$, an illustration of the difference between cross-source (blue) and within-source (red)
pairs for contrastive learning. Any-source uses
cross-source and within-source pairs.}
\label{fig:method_calda_contrastive}
\end{figure}

In the case of the first decision, we hypothesize that utilizing cross-source information to select example pairs can improve transfer. Prior MS-UDA methods \cite{wilson2020codats} ignore cross-source label information that can provide vital insights into how classes vary among domains (i.e., which aspects of the data truly indicate a different class label versus the same label from different domains). Utilizing this information can potentially improve transfer to the target domain. If our activity recognition class labels include ``walk'' and ``sleep'', we want the feature representations of two different walking people to be close, but the representations to be far apart for one walking person and one sleeping person. We use CALDA to investigate whether such cross-source information aids in transfer by explicitly making use of labeled data from each source domain as well as the differences in data distributions among the source domains. Furthermore, we compare this approach with a different instantiation of our framework that utilizes only labels within each source domain, thus explicitly ignoring cross-domain information. The differences between these approaches are illustrated in Figure~\ref{fig:method_calda_contrastive}.

For the second decision, we propose to utilize contrastive learning only across source domains rather than include pseudo-labeled target data that run the risk of being incorrectly labeled. Prior single-source domain adaptation methods have integrated contrastive losses \cite{park2020jcl,kang2019contrastive}. Because they utilize a single source domain with the target, they rely on pseudo-labeling the target domain data, creating difficult challenges when faced with large domain gaps \cite{choi2019pseudo}. Because CALDA employs more than one source domain, we can leverage a contrastive loss between source domains, thereby avoiding incorrectly pseudo-labeled target data.

For the third decision, we note that prior contrastive learning work has found selecting hard examples to yield improved performance \cite{schroff2015facenet,cai2020negativeselection}. However, recent theory postulates that hard examples do not need to be singled out - such examples already intrinsically yield a greater contribution to the contrastive loss than less-challenging examples \cite{khosla2020supervisedcontrastive}. We hypothesize that both random and hard sampling may offer improvements for multi-source domain adaptation, thus we evaluate both within CALDA.

CALDA integrates all of these components. As in prior work \cite{ganin2016jmlr,wilson2020codats}, we utilize a domain adversary that aligns feature representations across domains, yielding a domain-invariant feature extractor. Unlike prior approaches, we further utilize contrastive learning to pull examples from different source domains together in the feature space that have the same label while pushing apart examples from different domains that have different labels, though the choice of examples to pull and push depends on the three design decisions.

The key contribution of this paper is the development and evaluation of the CALDA framework for time-series MS-UDA. Specific contributions include:
\begin{itemize}
\item We improve upon existing time-series MS-UDA by leveraging cross-source domain labels via contrastive learning without requiring data augmentation or pseudo labeling.
\item We incorporate multiple contrastive learning strategies into CALDA to analyze the impact of design choices.
\item We offer an approach to time series MS-UDA that makes use of class distribution information where available through weak supervision.
\item We demonstrate performance improvements of CALDA over prior work with and without weak supervision. Improvements are shown for synthetic time series data and a variety of real-world time-series human activity recognition and electromyography datasets. These experiments aid in identifying the most promising CALDA instantiations, validating the importance of the adversary and unlabeled target data, and measuring the sensitivity of CALDA to noise within the weak supervision information.\footnote{Code and data is available at: \url{https://github.com/floft/calda}.}
\end{itemize}

\section{Related Work}

Here, we discuss related work on domain adaptation and contrastive learning in the context of time-series MS-UDA.

\subsection{Domain Adaptation}
Single-source domain adaptation methods abound \cite{wilson2020survey}, but little work studies multi-source domain adaptation. Zhao et al. \cite{zhao2018multisource} developed an adversarial method supporting multiple sources by including a binary domain classifier for each source. Ren et al. \cite{Ren22} align the sources, merging them into one domain that is aligned to the target.  Li et al. \cite{Li2021} also unify multiple source domains by updating model parameters to accommodate samples from each source. These approaches, however, do not take advantage of example similarities through contrastive learning to more effectively utilize source labels.
Xie et al. \cite{xie2017nips} propose a scalable method, only requiring one multi-class domain classifier. This approach is similar to the adversarial learning component of our framework.  As in our approach, Yadav et al. employ contrastive learning when combining multiple sources. Here, contrastive learning achieves higher intra-class compactness across domains, in the hope of yielding well-separated decision boundaries.
Yet, without models that are compatible with time-series, these approaches cannot be used for time series MS-UDA.

Limited research has investigated time-series domain adaptation, although these focus on a single source domain. Numerous approaches introduce explicit linear or nonlinear transformations to align the source and target spaces \cite{Ott2022,Shi2022,Cai2021,Hussein2022}. Some of these approaches, like CALDA, leverage an adversarial component. Liu et al. \cite{Liu2021} introduce a hybrid spectral kernel to characterize the non-stationary elements of the time series. A drawback is they do not make effective use of example similarity through contrastive learning. Purushotham et al. \cite{purushotham2017variational} developed a domain-adversarial method for single-source domain adaptation using a variational recurrent neural network (RNN) as the feature extractor. However, in our prior work \cite{wilson2020codats} we found that for both single-source and multi-source domain adaptation using a 1D convolutional neural network outperforms RNNs on a variety of time-series datasets. Thus,
we select this network architecture for our experiments.

Domain adaptation has also been studied specifically for electromyography (EMG)-based gesture recognition, mostly utilizing a different type of domain adaptation. Rather than UDA, several methods are proposed to improve supervised domain adaptation performance \cite{ketyko2019domain,ameri2019deep,cote2019emgmyo}, where some labeled target data are required. One method is developed for unsupervised domain adaptation \cite{du2017surface}, but they convert the EMG data to images followed by using adaptive batch normalization for domain adaptation \cite{li2018}. This approach makes the assumption that the domain differences are contained primarily within the network's normalization statistics and not the neural network layer weights. We do not require this assumption in our CALDA framework.

We uniquely incorporate weak supervision into domain adaptation. Weak supervision is inspired by the posterior regularization problem \cite{ganchev2010posterior}, but we consider this for domain adaptation. Along a similar vein, Jiang et al. \cite{jiang2018towards} use a related regularizer for the problem where label proportions are available for the source domains but not the target domain. Hu et al. \cite{hu2017learning} propose using a different form of weak supervision for human activity recognition from video data, using multiple incomplete or uncertain labels. Pathak et al. \cite{pathak2015iccv} develop a method for semantic segmentation using a weaker form of labeling than pixel-level labels. While we study weak supervision in a different context, the benefit of weak supervision in these other contexts in addition to the performance gains observed in our experiments suggests the general applicability of this idea. Additionally, prior weak supervision work fails to address the sensitivity of the weak supervision to noise \cite{wilson2020codats}, which we may expect with self-reported data. We analyze weak supervision in the presence of noise.

We select algorithm hyperparameters using cross-validation, but recent work offers alternative methods. Dinu et al. \cite{ICLR2023} proposed a theoretically-sound method by extending weighted least squares to deep neural networks for time series processing. The method offered by Saito et al. \cite{Saito21} relies on the intuition that a good classifier trained on source domain should embed close-by examples of the same class from the target close to form dense neighborhoods in the learned feature space. Density is measured by computing entropy of the similarity distribution between input examples. Furthermore, You et al. \cite{You19} select the best model by embedding feature representations into the validation procedure to obtain a unbiased estimation of the target risk.

\subsection{Contrastive Learning}
Our framework leverages the contrastive learning principle in addition to the adversarial learning from prior works. Early uses include clustering \cite{xing2002distance} and dimensionality reduction \cite{hadsell2006dimensionality}.
More recently, numerous research efforts have incorporated contrastive learning. These methods typically rely on data augmentation
to generate positives and negatives but sometimes use labels instead \cite{khosla2020supervisedcontrastive}. While such methods yield large gains in other contexts \cite{chen2020simple,he2020moco}, little work has explored contrastive learning for time series domain adaptation.

For single-source adaptation of image domains, prior methods \cite{park2020jcl,kang2019contrastive} consider a different contrastive loss. In addition to not utilizing an adversary, which we demonstrate is a vital component to our framework, these methods rely on data augmentation and pseudo labeling which may be problematic for time series. Data augmentation is standard and key to the success for many difficult image domain adaptations \cite{french2018iclr}, but is still being explored for time series \cite{iwana2021empirical} and is not required for CALDA. Second, prior methods perform contrastive learning on the combined (single) source domain and target domain. This critically depends on accurate target domain pseudo-labeling. However, pseudo-labeling remains a challenging problem that is particularly difficult when faced with the large domain gaps \cite{choi2019pseudo}, that frequently occur in time series data. Because we support multiple source domains, in CALDA we avoid pseudo-labeling and instead leverage the contrastive loss across source domains.
Other contrastive domain adaptation work focus on different problems: label-less transferable representation learning for image data \cite{thota2021contrastive} and image adaptation to a sequence of target domains \cite{su2020gradient}. As with the other prior work, these too rely on data augmentation \cite{thota2021contrastive,su2020gradient} and typically pseudo labeling \cite{su2020gradient}.

CALDA's final component is hard sampling, which is beneficial in some contrastive learning contexts. Schroff et al. \cite{schroff2015facenet} found it necessary for the triplet loss, a special case of contrastive learning
\cite{khosla2020supervisedcontrastive}. Similarly, Cai et al. \cite{cai2020negativeselection} found including the top  hard negatives to be both necessary and sufficient. While Khosla et al. \cite{khosla2020supervisedcontrastive} state that  sampling is not necessary since hard examples contribute more to the loss, this impact depends on having a large number of positives and negatives, which may not be optimal on all datasets, as demonstrated in our experimental results.

\section{Problem Setup}\label{sec:problem}
Here, we formalize Multi-Source Unsupervised Domain Adaptation (MS-UDA) without and with weak supervision.

\subsection{Multi-Source Unsupervised Domain Adaptation}
\label{section:MSUDA}
MS-UDA assumes that labeled data are available from multiple sources and unlabeled data are available from the target \cite{guo2018multi} to achieve the goal of creating a model that performs well on the target domain. Formally, given $n > 1$ source domain distributions $\mathcal{D}_{S_i}$ for $i \in \{ 1, 2, \dots, n \}$ and a target domain distribution $\mathcal{D}_T$, we draw $s_i$ labeled training examples i.i.d. from each source distribution $\mathcal{D}_{S_i}$ and $t_{train}$ unlabeled training instances i.i.d. from the marginal distribution $\mathcal{D}_T^X$:

\begin{equation}
    S_i = \{ (\textbf{x}_j, y_j) \}^{s_i}_{j=1} \sim \mathcal{D}_{S_i} \quad \forall i \in \{ 1, 2, \dots, n \}
\end{equation}
\begin{equation}
    T_{train} = \{ (\textbf{x}_j) \}^{t_{train}}_{j=1} \sim \mathcal{D}_T^X
\end{equation}

Here, each domain is distributed over the space $X \times Y$, where $X$ is the input data space and $Y$ is the label space $Y = \{ 1, 2, \dots, L \}$ for $L$ classification labels. After training a MS-UDA model $f: X \rightarrow Y$ using $S_i$ and $T_{train}$, we test the model using a holdout set of $t_{test}$ labeled testing examples (input and ground-truth label pairs) drawn i.i.d. from the target distribution $\mathcal{D}_T$:
\begin{equation}
  T_{test} = \{ (\textbf{x}_j, y_j) \}^{t_{test}}_{j=1} \sim \mathcal{D}_T
\end{equation}

In the case of time series domain adaptation, $X = [X^1, X^2, \dots, X^K]$ for $K$ time series variables or channels. Each variable $X^i$ for $i \in \{ 1, 2, \dots, K \}$ consists of a time series $X^i = [x_1, x_2, \dots, x_H]$ containing a sequence of real values observed at equally-spaced time steps $1, 2, \dots H$  \cite{fawaz2019review}.

\subsection{MS-UDA with Weak Supervision}

When MS-UDA is guided by weak supervision \cite{wilson2020codats}, the target-domain label proportions are additionally available during training, which we can utilize to guide the neural network's representation. Formally, these proportions represent $P(Y=y)$ for the target domain, i.e., the probability $p_y$ that each example will have label $y \in \{ 1, 2, \dots, L \}$:
\begin{equation}\label{eq:calda_labelproportions}
    Y_{true}(y) = P(Y=y) = p_y
\end{equation}

\begin{figure}
  \centering
  \includegraphics[width=1.0\linewidth]{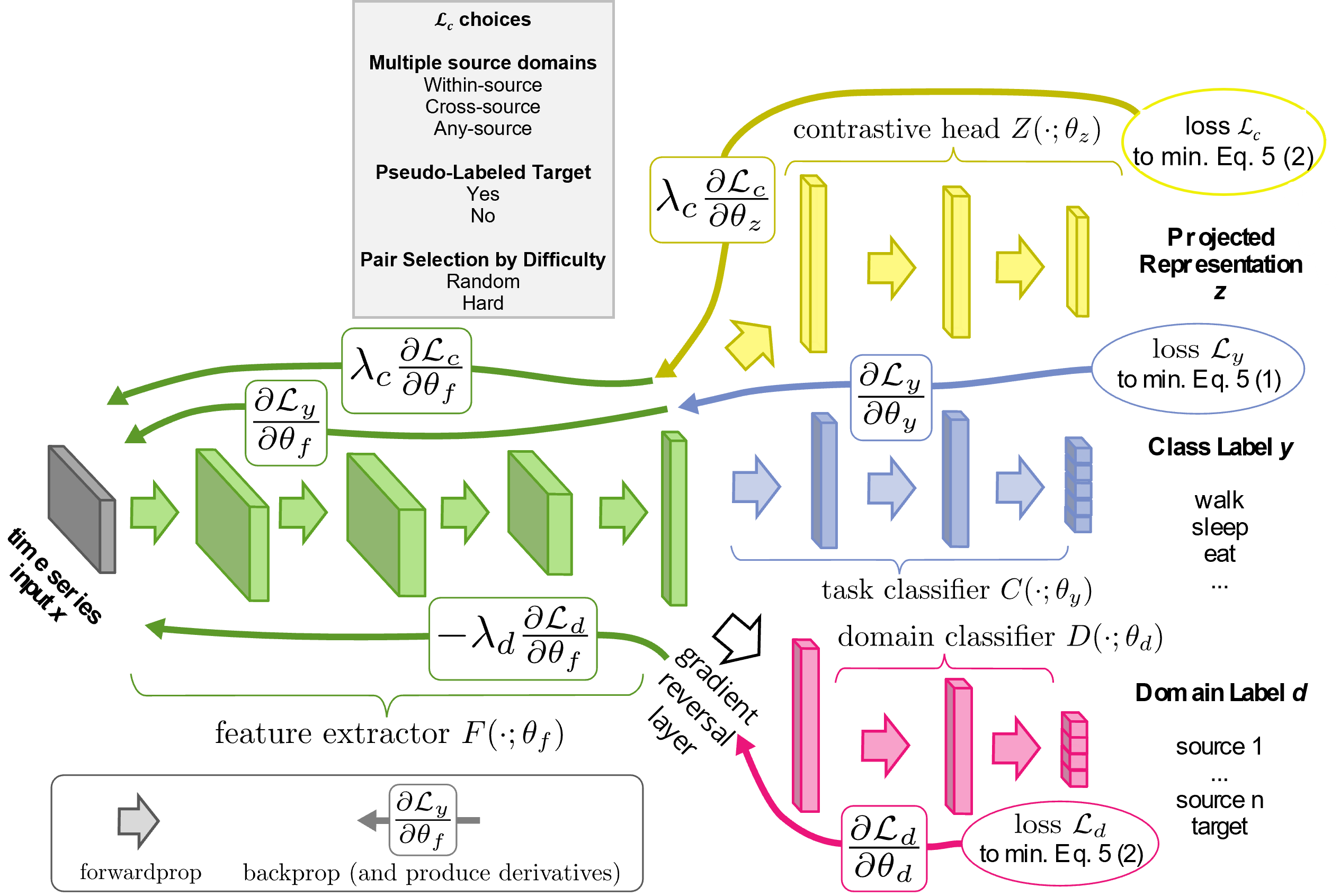}
  \caption[The CALDA MS-UDA framework]{CALDA incorporates adversarial learning via a domain classifier and contrastive learning via a contrastive loss on an additional contrastive head in the network. Through the CALDA instantiations, we determine how to best utilize contrastive learning for MS-UDA.}
  \label{fig:method_calda_components}
\end{figure}

\section{CALDA Framework}

We introduce CALDA, a MS-UDA framework that blends adversarial learning with contrastive learning. First, we motivate CALDA from domain adaptation theory. Second, we describe the key components: source domain error minimization, adversarial learning, and contrastive learning. Finally, we describe framework alternatives to investigate how to best construct the example sets used in contrastive loss.

\subsection{Theoretical Motivation}
Zhao et al. \cite{zhao2018multisource} offer an error bound for multi-source domain adaptation. Given a hypothesis space $\mathcal{H}$ with VC-dimension $v$, $n$ source domains, empirical risk $\hat{\epsilon}_{S_i}(h)$ of the hypothesis on source domain $S_i$ for $i \in \{ 1, 2, \dots, n \}$, empirical source distributions $\hat{\mathcal{D}}_{S_i}$ for $i \in \{ 1, 2, \dots, n \}$ generated by $m$ labeled samples from each source domain, empirical target distribution $\hat{\mathcal{D}}_T$ generated by $mn$ unlabeled samples from the target domain, optimal joint hypothesis error $\lambda_\alpha$ on a mixture of source domains $\sum_{i \in [n]} \alpha_i S_i$, and target domain $T$ (average case if $\alpha_i = 1/n$ $\forall i \in \{ 1, 2, \dots n \}$), the target classification error bound $\epsilon_T(h)$ with probability at least $1 - \delta$ for all $h \in \mathcal{H}$ can be expressed as:

\scriptsize
\begin{equation}\label{eq:theory}
    \begin{aligned}
        \epsilon_T(h) &
        \leq \sum_{i=1}^n \alpha_i \Big( \overbrace{\hat{\epsilon}_{S_i}(h)}^{\text{(1) source errors}} + \underbrace{\frac{1}{2} d_{\mathcal{H} \Delta \mathcal{H}}(\hat{\mathcal{D}}_T; \hat{\mathcal{D}}_{S_i})}_{\text{(2) divergences}} \Big) \\
        & + \overbrace{\lambda_\alpha}^{\text{(3) opt. joint hyp.}} + \underbrace{O \left( \sqrt{\frac{1}{nm} \left( \log \frac{1}{\delta} + v \log \frac{nm}{v} \right)} \right)}_{\text{(4) due to finite samples}}
    \end{aligned}
\end{equation}
\normalsize

In Equation~\ref{eq:theory}, term (1) is the sum of source domain errors, (2) is the sum of the divergences between each source domain and the target, (3) is the optimal joint hypothesis on the mixture of source domains and the target domain, and (4) addresses the finite sample sizes. Note that the first two terms are the most relevant for informing multi-source domain adaptation methods since they can be optimized. In contrast, given a hypothesis space (e.g., a neural network of a particular size and architecture), (3) is fixed. Similarly, (4) regards finite samples from all domains, which depends on the number of samples and for a given dataset cannot increase.

We introduce CALDA to minimize this error bound as illustrated in Figure~\ref{fig:method_calda_components}. First, we train a {\em Task Classifier} to correctly predict the labeled data from the source domains, thus minimizing (1). To minimize (2), we better align domains based on adversarial learning and contrastive learning. As in prior works, we use feature-level domain invariance via domain adversarial training to align the sources and unlabeled data from the target domain. We train a {\em Domain Classifier} to predict which domain originated a representation while simultaneously training the {\em Feature Extractor} to generate domain-invariant representations that fool the Domain Classifier. Additionally, we propose a supervised contrastive loss to align the representations of same-label examples among the multiple source domains. The new loss aids in determining which aspects of the data correspond to differences in the class label (the primary concern) versus differences in the domain (e.g., person) where the data originated (which can be ignored). The new loss definition leverages both the labeled source domain data and cross-source information. This contrastive loss is applied to an additional {\em Contrastive Head} in the model. To address term (3), we consider an adequately-large hypothesis space by using a neural network of sufficient size and incorporating an architecture previously shown to handle time-series data \cite{wilson2020codats,wang2017strongbaseline}.

\subsection{Adaptation Components}

\begin{figure}
  \centering
  \includegraphics[width=0.7\linewidth]{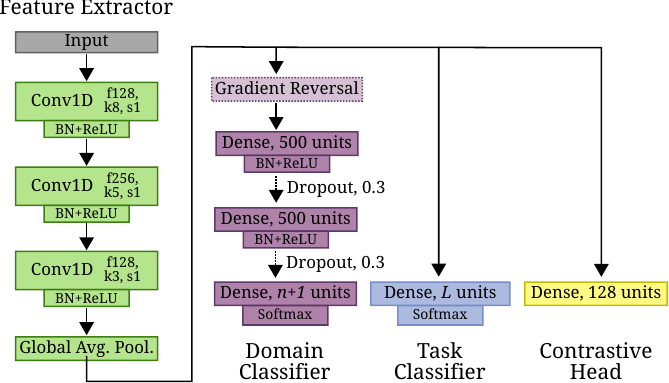}
  \caption[The CALDA model architecture]{The CALDA model architecture consists of a CNN task classifier, multi-layer perceptron domain classifier with global average pooling, and a contrastive head.}
  \label{fig:model}
\end{figure}

CALDA's adaptation architecture is shown in Figure~\ref{fig:model}. The architecture includes a feature extractor, task classifier, domain classifier, and contrastive head. As illustrated in the figure, the feature extractor consists of a fully convolutional network, where the dense last layer acts as the task classifier. The domain classifier, consisting of a multi-layer perceptron, performs the adversary role during training. To handle variable-length time series data, the CNN includes a global average pooling layer. Finally, CALDA includes an additional contrastive head to support the contrastive loss.

We describe source domain errors and feature-level domain invariance before moving onto our novel contrastive loss for multi-source domain adaptation, optionally with weak supervision, and the corresponding design choices.

\subsubsection{Minimize Source Domain Errors}
We minimize classification error on the source domains by feeding the outputs of feature extractor $F(\cdot; \theta_f)$ to a task classifier $C(\cdot; \theta_c)$ having a softmax output. Then, we update the parameters $\theta_f$ and $\theta_c$ to minimize a categorical cross-entropy loss $\mathcal{L}_y(y,p)$ using one-hot encoded true label $y$ and softmax probabilities $p$. To handle multiple sources, we compute this loss over a mini-batch of examples drawn from each of the source domain distributions $\mathcal{D}_{S_i}$ for $i \in \{ 1, 2, \dots n \}$:
\begin{equation}\label{eq:task}
  \argmin_{\theta_f,\theta_c} \sum_{i=1}^n \mathop{\mathbb{E}}_{(x, y) \sim \mathcal{D}_{S_i}} \left[ \mathcal{L}_y(y, C(F(x))) \right]
\end{equation}

We employ the categorical cross-entropy loss, where $y_i$ and $p_i$ represent the $i$th components of $y$'s one-hot encoding and the softmax probability output vector, respectively:
\begin{equation}
  \mathcal{L}_y(y,p) = - \sum_{i=1}^L y_i \log p_i
\end{equation}

\subsubsection{Adversarial Learning}
If we rely on only minimizing source domain error, we will obtain a classifier that likely does not transfer well to the target domain. One reason for this is that the extractor's selected feature representations may differ widely between source domains and the target domain. To remedy this problem, we invite a domain adversary to produce feature-level domain invariance. In other words, we align the feature extractor's outputs across domains. We achieve alignment by training a domain classifier (the ``adversary'') to correctly predict each example's domain labels (i.e., predict which domain each example originated from) while simultaneously training the feature extractor to make the domain classifier predict that the example is from a different domain.

We define a multi-class domain classifier with a softmax output as the adversary \cite{wilson2020codats}. The domain classifier $D(\cdot; \theta_d)$ follows the feature extractor $F$ in the network. However, we place a gradient reversal layer $\mathcal{R}(\cdot)$ between $F$ and $D$, which multiplies the gradient during backpropagation by a negative constant $-\lambda_{d}$, yielding adversarial training \cite{ganin2016jmlr}. Given domain labels $d \in \{ 0, 1, \dots, n \}$, that map target examples to label $d_T=0$ and source $i$ examples to label $d_{S_i} = i$ for $i \in \{ 1, 2, \dots n \}$, we update the model parameters $\theta_f$ and $\theta_d$:

\small
\begin{equation}\label{eq:domain}
  \begin{aligned}
    \argmin_{\theta_f,\theta_d}
      & \sum_{i=1}^n \mathop{\mathbb{E}}_{(x, y) \sim \mathcal{D}_{S_i}} \left[ \mathcal{L}_d(d_{S_i}, D(\mathcal{R}(F(x)))) \right] \\
      & + \mathop{\mathbb{E}}_{x \sim \mathcal{D}_T^X} \left[ \mathcal{L}_d(d_T, D(\mathcal{R}(F(x)))) \right] \\
  \end{aligned}
\end{equation}
\normalsize

This objective incorporates a categorical cross-entropy loss $\mathcal{L}_d$ similar to $\mathcal{L}_y$ that uses {\em domain labels} instead of class labels. Given the one-hot encoded representation of the true domain label $d$ and the domain classifier's softmax probability output vector $p$, we compute the loss:
\begin{equation}
  \mathcal{L}_d(d,p) = - \sum_{i=1}^n d_i \log p_i
\end{equation}

\subsubsection{Contrastive Learning}
The above domain invariance adversarial loss does not leverage labeled data from the source domains. Prior work \cite{wilson2020codats} only indirectly leverages labeled source domain data through jointly training the adversarial loss and task classifier loss on the labeled source domains. To better utilize source labels, we propose employing a supervised contrastive loss \cite{khosla2020supervisedcontrastive} to pull same-labeled examples together in the embedding space and push apart different-labeled examples, thereby making use of both positive and negative cross-source label information.

While the exact details vary based on the design decisions we will discuss in the next section, in general, contrastive learning has two roles: (1) pull same-label examples together and (2) push different-label examples apart. This process operates on pairs of representations $z$ of examples $(z_1, z_2)$. We call the first $z_1$ the ``query'' or ``anchor'', i.e., $z_1 = q$, where $q \in Q$ is drawn from the set of all example representations. To pull examples together, we create a pair $(q, p)$, where the ``positive'' $p \in P$ is drawn from the set of all example representations having the \textit{same label} as $q$. To push examples apart, we create another pair $(q, n)$, where ``negative'' $n \in N$ is drawn from the set of example representations that have a \textit{different label} than $q$. CALDA allows additional constraints to be placed on how positives and negatives are selected, such as selecting examples from the same domain, a different domain, or any domain. Figure~\ref{fig:method_calda_contrastive} illustrates one query positive pair $(q, p)$ and negative pair $(q, n)$ for the cross-source and within-source cases. We may create additional positive and negative pairs for each query similarly.

We propose using a {\em supervised contrastive loss} based on a multiple-positive InfoNCE loss \cite{oord2018infonce,khosla2020supervisedcontrastive}. Given the projected representation $z$ of a query, the corresponding positive and negative sets $P$ and $N$, a temperature hyperparameter $\tau$, and cosine similarity $\text{sim}(z_1, z_2) = \frac{z_1^T z_2}{\|z_1\|\|z_2\|}$, we obtain:
\begin{equation}\label{eq:contrastive}
  \begin{aligned}
    & \mathcal{L}_c(z, P, N) = \\
    & \frac{1}{|P|} \sum_{z_p \in P} \left[ - \log \left( \frac{\exp\left({\frac{\text{sim}\left(z, z_p\right)}{\tau}}\right)}{\sum_{z_k \in N \cup \{z_p\}} \exp\left({\frac{\text{sim}\left(z, z_k\right)}{\tau}}\right)} \right) \right]
  \end{aligned}
\end{equation}

Conceptually, for a given query, Equation~\ref{eq:contrastive} sums over each positive and normalizes by the number of positives. Inside of this sum, we compute what is mathematically equivalent to a log loss for a softmax-based classifier that classifies the query as the positive \cite{he2020moco}. The denominator sums over both the positive and also all the negatives corresponding to the query. Note, alternatively, the sum over the positive set could be moved inside the log, but keeping this sum outside has been found to perform better in prior works using InfoNCE \cite{khosla2020supervisedcontrastive}.

Finally, we update weights by summing over the queries for each source domain and normalizing by the number of queries. In some framework instantiations, we similarly compute this loss over queries from the pseudo-labeled target domain data. Formally, given the source domain queries $Q_{S_i}$, the pseudo-labeled target domain queries $Q_T$, positive and negative sets $P_{d,y}$ and $N_{d,y}$ (construction depends on the instantiation of our method, discussed next), and the indicator function $\textbf{1}$, we can update the model parameters $\theta_f$ and $\theta_z$:
\scriptsize
\begin{equation}\label{eq:contrastive_objective}
  \begin{aligned}
    \argmin_{\theta_f,\theta_z} \sum_{i=1}^n & \left[ \frac{1}{|Q_{S_i}|} \sum_{(z_q, y_q) \in Q_{S_i}} \mathcal{L}_c(z_q, P_{d_{S_i}, y_q}, N_{d_{S_i}, y_q}) \right] \\
    + \textbf{1}_{PL=True} & \frac{1}{|Q_T|} \sum_{(z_q, \hat{y}_q) \in Q_T} \mathcal{L}_c(z_q, P_{d_T, \hat{y}_q}, N_{d_T, \hat{y}_q})
  \end{aligned}
\end{equation}
\normalsize

\subsubsection{Total Loss and Weak Supervision Regularizer}
We jointly train each of these three adaptation components. Thus, the total loss that we minimize during training is a sum of each loss: source domain errors from Equation~\ref{eq:task}, adversarial learning from Equation~\ref{eq:domain}, and contrastive learning from Equation~\ref{eq:contrastive_objective}. We further add a weighting parameter $\lambda_c$ for the contrastive loss and note that the $\lambda_d$ multiplier included in the gradient reversal layer $\mathcal{R}(\cdot)$ can be used as a weighting parameter for the adversarial loss.

Additionally, for the problem of MS-UDA with weak supervision, we include the weak supervision regularization term described in our prior work \cite{wilson2020codats}. While individual labels for the unlabeled target domain data are unknown, this KL-divergence regularization term guides training toward learning model parameters that produce a class label distribution approximately matching the given label proportions on the unlabeled target data. This allows us to leverage target-domain label distribution information, if available.

\begin{equation}\label{eq:calda_ws}
  \argmin_{\theta_f,\theta_c} \left[ \kld*{Y_{true}}{\mathbb{E}_{x \sim \mathcal{D}_T^X} \big[ C(F(x)) \big]} \right]
\end{equation}

\subsection{Design Decisions for Contrastive Learning}

The contrastive losses used in Equation~\ref{eq:contrastive_objective} require positive and negative pairs for each query. CALDA supports multiple options for selecting these pairs.
Here, we formalize the CALDA instantiations along the dimensions of 1) how to select example pairs across multiple domains, 2) whether to include a pseudo-labeled target domain in contrastive learning, and 3) whether to select examples randomly or based on difficulty.

\subsubsection{Multiple Source Domains}
When selecting pairs of examples for MS-UDA, we may choose to select two examples within a single domain, from two different domains, or a combination. We term these variations {\em Within-Source}, {\em Cross-Source}, and {\em Any-Source} Label-Contrastive learning. Note that similar terms apply if including the pseudo-labeled target domain. Recall that the motivation behind contrastive-learning MS-UDA is to leverage cross-source information. Because cross-source information is excluded in the Within-Source case, we hypothesize that Within-Source will perform poorly, whereas Any-Source and Cross-Source, which leverage the cross-source information, will yield improved results.

Formally, we define the sets of queries, positives, and negatives for each of these cases using set-builder notation. To simplify constructing these sets, we first create the auxiliary set $K$, which contains input $x$, class label $y$ (or in the case of the target domain, the pseudo-label $\hat{y}$), and domain label $d$ for all examples. Given a set of labeled examples $S_i \sim \mathcal{D}_{S_i}$ from each source domain $i \in \{ 1, 2, \dots, n \}$, a set of unlabeled instances from the target domain $T_{train} \sim \mathcal{D}_T$, and a \textit{projected representation} defined as the feature-level representation passed through an additional contrastive head $Z(\cdot; \theta_z)$ in the model $z = Z(F(x))$ (e.g., an additional fully-connected layer), we define a set $K_S$ including both the $(x,y)$ pair and the domain label $d = d_{S_i}$ (as defined in the previous section) of all source domains, a set $K_T$ including both the $(x,\hat{y})$ pseudo-labeled pair and the domain label $d = d_T$ of the target domain, and set $K$, which is the union of $K_S$ and $K_T$:

\small
\begin{equation}
  K_S = \{ (x, y, d) \mid (x, y) \in S_i, i \in \{ 1, 2, \dots, n \}, d = d_{S_i} \}
\end{equation}
\begin{equation}
  K_T = \{ (x, \hat{y}, d) \mid x \in T_{train}, \hat{y} = \argmax C(F(x)), d = d_T \}
\end{equation}
\begin{equation}
  K = K_S \cup K_T
\end{equation}
\normalsize
Using $K$, we define the query set $Q_{S_i}$ for each source domain and the query set $Q_T$ for the target domain:
\small
\begin{equation}
  Q_{S_i} = \{ (z, y) \mid (x, y, d) \in K, z = Z(F(x)), d = d_{S_i} \}
\end{equation}
\begin{equation}
  Q_T = \{ (z, \hat{y}) \mid (x, \hat{y}, d) \in K, z = Z(F(x)), d = d_T \}
\end{equation}
\normalsize

Next, we define the positive and negative sets for each framework instantiation.

{\bf(a)  Within-Source Label-Contrastive learning:} Positives for each query are selected  from the \textit{same domain} as the query with the \textit{same label}. Negatives are selected that from the \textit{same domain} as the query with a \textit{different label}. Formally, we define the positive and negative sets $P_{d_q,y_q}$ and $N_{d_q,y_q}$ for each query of domain $d_q$ and label $y_q$ as follows:

\small
\begin{equation}
  P_{d_q, y_q} = \{ z \mid (x, y, d) \in K, z = Z(F(x)), d = d_q, y = y_q \}
\end{equation}
\begin{equation}
  N_{d_q, y_q} = \{ z \mid (x, y, d) \in K, z = Z(F(x)), d = d_q, y \neq y_q \}
\end{equation}
\normalsize

{\bf (b) Any-Source Label-Contrastive learning:} Positives for each query are selected having the \textit{same label} and coming from \textit{any domain}. Negatives are selected with a \textit{different label} and from \textit{any domain}:

\small
\begin{equation}
  P_{d_q, y_q} = \{ z \mid (x, y, d) \in K, z = Z(F(x)), y = y_q \}
\end{equation}
\begin{equation}
  N_{d_q, y_q} = \{ z \mid (x, y, d) \in K, z = Z(F(x)), y \neq y_q \}
\end{equation}
\normalsize

{\bf (c) Cross-Source Label-Contrastive learning:} Positives for each query are selected from a \textit{different domain} with the \textit{same label}. Negatives are selected from a \textit{different domain} with a \textit{different label}:

\small
\begin{equation}
  P_{d_q, y_q} = \{ z \mid (x, y, d) \in K, z = Z(F(x)), d \neq d_q, y = y_q \}
\end{equation}
\begin{equation}
  N_{d_q, y_q} = \{ z \mid (x, y, d) \in K, z = Z(F(x)), d \neq d_q, y \neq y_q \}
\end{equation}
\normalsize

Note that these cases are distinguished based on whether $d = d_q$ (Within-Source), $d \neq d_q$ (Cross-Source), or there is no constraint (Any-Source).

\subsubsection{Pseudo-Labeled Target Domain}
Prior contrastive learning work for single-source domain adaptation utilizes a supervised contrastive loss on the combined single source domain and the target domain. However, since this loss depends on labels, such methods require pseudo-labeling the target domain data. The methods rely on the classifier producing correct class labels, which can then be used in the supervised contrastive loss. Unfortunately, pseudo-labeling the target domain is a challenging problem, and classification errors are likely \cite{choi2019pseudo}. If the pseudo-labels are incorrect, then this may hurt contrastive learning performance. Because US-MDA utilizes multiple domains, instead of performing contrastive learning between the source and target domains, we may perform contrastive learning among the source domains, which we may improve performance. We include whether to utilize pseudo-labeled target domain data during training as an additional CALDA dimension. If pseudo-labeled target domain data is included during contrastive learning, $PL = True$ in the contrastive learning objective (Equation~\ref{eq:contrastive_objective}), otherwise $PL = False$.

\subsubsection{Pair Selection by Difficulty}
Outside of domain adaptation, selecting \textit{hard} examples has been found beneficial for contrastive learning \cite{schroff2015facenet,cai2020negativeselection}. However, recent theoretical work suggests that hard examples implicitly contribute more to the loss, thus mitigating the need for explicitly selecting hard examples in contrastive learning \cite{khosla2020supervisedcontrastive}. To determine if explicitly selecting hard examples is beneficial in multi-source domain adaptation, we propose a method for hard sampling in CALDA and compare it with random sampling -- the final dimension of our CALDA framework. For brevity, we give the equations for Cross-Source Label-Contrastive learning, but the other variations can be constructed by changing the domain constraint of each set.

For \textit{hard sampling}, we select a subset of hard positive and negative examples. This necessitates that we define ``hard examples.'' In each case, the domain constraint is the same for both positives and negatives, thus the key difference is whether they have the same label as the query or not. The examples that would most help the model learn this decision boundary are those that are predicted to be on the wrong side. Thus, we select \textit{hard} examples as examples that are currently predicted to be on the wrong side of this decision boundary. We define \textit{hard positives} as examples with the query's label but with a different predicted label (with respect to the current model predictions) and \textit{hard negatives} as examples of a class other than the query's label but that are predicted to have the query's label. Both are from a different domain than the query (in the Cross-Source case). Focusing on the Cross-Source Label-Contrastive case, for a query with domain $d_q$ and true label $y_q$ and the current model prediction $\hat{y} = \argmax C(F(x))$, we define hard positive and negative sets $\bar{P}$ and $\bar{N}$:
\begin{equation}\label{eq:hard_p}
  \begin{aligned}
    \bar{P}_{d_q,y_q} = \{ z \mid
      & (x, y, d) \in K, z = Z(F(x)), \\
      & d \neq d_q, y = y_q, \hat{y} \neq y_q \}
  \end{aligned}
\end{equation}
\begin{equation}\label{eq:hard_n}
  \begin{aligned}
    \bar{N}_{d_q,y_q} = \{ z \mid
      & (x, y, d) \in K, z = Z(F(x)), \\
      & d \neq d_q, y \neq y_q, \hat{y} = y_q \}
  \end{aligned}
\end{equation}

However, there is no guarantee there will always be positives and negatives that are misclassified. For example, the task classifier likely makes accurate predictions later on during training. Instead, we propose using a relaxed version of hardness: take the top-$k_1$ hardest positives and top-$k_2$ hardest negatives in terms of a softmax cross-entropy loss. To obtain a relaxation of $\hat{y} \neq y_q$ for the positives, we can find positive examples with a \textit{high} task classification loss for that example via $\mathcal{L}_y$. This is because a positive is defined as having the same label as the query, so having a high task loss for the positive corresponds to being on the wrong side of the positive-negative decision boundary. To obtain a relaxation of $\hat{y} = y_q$ for the negatives, we can find negative examples with a \textit{low} task classification loss where we replace the true class label $y$ with the query's class label $y_q$. This finds the negatives that are most easily misclassified as having the query's class label, i.e., those on the wrong side of the decision boundary.

Thus, we define the relaxed hard positive and negative sets $\tilde{P}$ and $\tilde{N}$ in terms of the softmax-based cross-entropy loss $\mathcal{L}_y$, with loss thresholds $h_p$ and $h_n$ chosen such that we have $k_1$ positives and $k_2$ negatives (i.e, $|P| = k_1$ and $|N| = k_2$):

\small
\begin{equation}\label{eq:relaxed_p}
  \begin{aligned}
    \tilde{P}_{d_q,y_q} = \{ z \mid
      & (x, y, d) \in K, z = Z(F(x)), d \neq d_q, y = y_q, \\
      & \mathcal{L}_y(y, C(F(x))) > h_p \}
  \end{aligned}
\end{equation}
\begin{equation}\label{eq:relaxed_n}
  \begin{aligned}
    \tilde{N}_{d_q,y_q} = \{ z \mid
      & (x, y, d) \in K, z = Z(F(x)), d \neq d_q, y \neq y_q, \\
      & \mathcal{L}_y(y_q, C(F(x))) < h_n \}
  \end{aligned}
\end{equation}
\normalsize

The contrastive weight update in Equation~\ref{eq:contrastive_objective} can now be adjusted to use the relaxed hard positive and negative sets $\tilde{P}$ and $\tilde{N}$. As an alternative to hard sampling, we may instead use \textit{random sampling}, to select a random subset of positives and  negatives that pair with each query.

\section{Experimental Validation}

We validate our hypothesis that CALDA will improve time-series MS-UDA through contrastive learning based on experimental analysis. We also apply CALDA to synthetic and real-world datasets to address the three design decisions, with and without weak supervision. Finally, we validate the impact of the adversary and unlabeled target domain data.

\subsection{Datasets}

We construct several synthetic time series that vary in complexity to aid in comparing alternative adaptation frameworks. He et al. \cite{He2023} observe that adapting models to new domains becomes more challenging as the data become more complex. We therefore generate synthetic data of varying complexity to examine corresponding differences in adaptation performance.
In the first scenario (SW), we generate a 2D normal distribution for each domain that represents a sum of two sine waves with frequencies $f_{i,1}$ and $f_{i,2}$ (Hz), obtaining a time series example $(x_i, y_i)$, where $x_i$ is a vector. In this sine wave scenario, inter-domain or intra-domain separation is created through translating or rotating the class distributions by a fixed amount (examples are illustrated in the Supplementary Material). In the second scenario, the synthetic data more closely resemble the complexities found in the real-world datasets. We generate multivariate (9 dimensions) data created from a mixture of Gaussians. Separation between domains is created by varying the Gaussian parameters for each dimension. We repeat this scenario for 1, 2, and 3 Gaussians (scenarios 1GMM, 2GMM, and 3GMM) to vary the corresponding complexity of the data distributions and domain-invariant features that must be learned.
Note that for both scenarios, as the number of source domains $n$ increases, the source domains cover a larger region of the space compared to the target domain.

\begin{table*}
\centering
\caption[Ablation study of CALDA instantiations that include the target-domain data via pseudo labeling]{Ablation study of CALDA instantiations that include the target-domain data via pseudo labeling.}
\label{table:ablation_pseudo}
\begin{scriptsize}
{\renewcommand{\arraystretch}{1.4}
\begin{tabular}{ccccccc}
\toprule
Dataset & \textit{CALDA-In,R,P} & \textit{CALDA-In,H,P} & \textit{CALDA-Any,R,P} & \textit{CALDA-Any,H,P} & \textit{CALDA-XS,R,P} & \textit{CALDA-XS,H,P} \\
\midrule
UCI HAR & 92.3 $\pm$ 2.5 & 92.0 $\pm$ 3.0 & 92.6 $\pm$ 2.6 & 91.9 $\pm$ 3.3 & 92.1 $\pm$ 3.0 & 92.4 $\pm$ 3.3 \\
UCI HHAR & 89.2 $\pm$ 4.2 & 89.0 $\pm$ 4.5 & 88.8 $\pm$ 4.7 & 86.2 $\pm$ 5.8 & 89.4 $\pm$ 4.3 & 87.4 $\pm$ 5.6 \\
WISDM AR & 72.5 $\pm$ 9.1 & 71.1 $\pm$ 8.0 & 74.7 $\pm$ 8.3 & 71.3 $\pm$ 9.1 & 74.5 $\pm$ 8.4 & 73.5 $\pm$ 8.3 \\
WISDM AT & 68.6 $\pm$ 8.7 & 68.5 $\pm$ 7.2 & 61.6 $\pm$ 12.9 & 63.3 $\pm$ 9.8 & 62.0 $\pm$ 10.9 & 60.2 $\pm$ 12.3 \\
Myo EMG & 83.4 $\pm$ 5.5 & 82.4 $\pm$ 5.4 & 83.7 $\pm$ 5.5 & 82.9 $\pm$ 6.2 & 84.2 $\pm$ 5.3 & 82.0 $\pm$ 6.4 \\
NinaPro Myo & 52.0 $\pm$ 5.3 & 52.6 $\pm$ 4.5 & 51.0 $\pm$ 4.6 & 52.4 $\pm$ 4.1 & 51.5 $\pm$ 5.3 & 52.3 $\pm$ 4.6 \\
\hline
Average & 77.2 $\pm$ 5.9 & 76.7 $\pm$ 5.5 & 76.2 $\pm$ 6.5 & 75.4 $\pm$ 6.5 & 76.5 $\pm$ 6.2 & 75.4 $\pm$ 6.8 \\
\bottomrule
\end{tabular}

}
\end{scriptsize}
\end{table*}

\begin{table*}
\centering
\caption[Ablation study comparing hard and random sampling for each CALDA instantiation]{Ablation study comparing hard and random sampling for each CALDA instantiation. Bold denotes highest accuracy in each row.}
\label{table:ablation_selection}
\begin{scriptsize}
{\renewcommand{\arraystretch}{1.4}
\begin{tabular}{ccccccc}
\toprule
Dataset & \textit{CALDA-In,R} & \textit{CALDA-In,H} & \textit{CALDA-Any,R} & \textit{CALDA-Any,H} & \textit{CALDA-XS,R} & \textit{CALDA-XS,H} \\
\midrule
UCI HAR & 93.5 $\pm$ 2.0 & 93.6 $\pm$ 2.2 & 93.1 $\pm$ 2.3 & \textbf{93.7 $\pm$ 2.1} & 93.4 $\pm$ 2.1 & 93.4 $\pm$ 2.5 \\
UCI HHAR & 89.3 $\pm$ 3.9 & 89.4 $\pm$ 3.9 & \textbf{89.8 $\pm$ 3.7} & 88.7 $\pm$ 4.6 & \textbf{89.8 $\pm$ 3.9} & 89.4 $\pm$ 4.0 \\
WISDM AR & 79.0 $\pm$ 6.9 & 79.4 $\pm$ 7.7 & 80.2 $\pm$ 7.1 & 80.3 $\pm$ 6.9 & 80.0 $\pm$ 6.3 & \textbf{81.4 $\pm$ 7.9} \\
WISDM AT & 71.4 $\pm$ 8.2 & 71.0 $\pm$ 8.3 & \textbf{72.1 $\pm$ 7.5} & 71.2 $\pm$ 8.4 & 70.7 $\pm$ 6.5 & 71.0 $\pm$ 8.6 \\
Myo EMG & 83.0 $\pm$ 5.3 & 82.7 $\pm$ 6.1 & \textbf{83.8 $\pm$ 5.6} & 83.6 $\pm$ 5.4 & 83.4 $\pm$ 5.6 & 83.3 $\pm$ 5.3 \\
NinaPro Myo & 57.3 $\pm$ 3.6 & 56.9 $\pm$ 3.6 & 57.7 $\pm$ 3.8 & 55.9 $\pm$ 4.2 & \textbf{58.0 $\pm$ 3.8} & 56.0 $\pm$ 3.9 \\
\hline
Synth InterT 0 & 93.7 $\pm$ 0.2 & 93.7 $\pm$ 0.1 & 93.8 $\pm$ 0.2 & 93.7 $\pm$ 0.2 & 93.7 $\pm$ 0.3 & \textbf{93.9 $\pm$ 0.1} \\
Synth InterR 0 & \textbf{94.2 $\pm$ 0.2} & \textbf{94.2 $\pm$ 0.1} & 94.1 $\pm$ 0.1 & 94.1 $\pm$ 0.2 & 94.1 $\pm$ 0.1 & 94.0 $\pm$ 0.2 \\
Synth IntraT 0 & 93.7 $\pm$ 0.2 & 93.8 $\pm$ 0.2 & 93.8 $\pm$ 0.3 & \textbf{93.9 $\pm$ 0.2} & 93.7 $\pm$ 0.3 & 93.6 $\pm$ 0.3 \\
Synth IntraR 0 & \textbf{94.1 $\pm$ 0.2} & \textbf{94.1 $\pm$ 0.2} & \textbf{94.1 $\pm$ 0.2} & 94.0 $\pm$ 0.1 & \textbf{94.1 $\pm$ 0.2} & 93.9 $\pm$ 0.2 \\
\hline
Synth InterT 10 & 69.8 $\pm$ 17.0 & \textbf{70.8 $\pm$ 15.8} & 69.4 $\pm$ 17.0 & 70.7 $\pm$ 15.0 & 68.4 $\pm$ 14.2 & 70.5 $\pm$ 14.5 \\
Synth InterR 1.0 & 67.5 $\pm$ 12.5 & 69.9 $\pm$ 12.6 & 63.4 $\pm$ 10.2 & \textbf{77.4 $\pm$ 8.1} & 63.8 $\pm$ 10.3 & 76.3 $\pm$ 8.9 \\
Synth IntraT 10 & 74.9 $\pm$ 6.4 & 73.6 $\pm$ 6.6 & 75.6 $\pm$ 8.4 & 75.9 $\pm$ 8.9 & \textbf{77.9 $\pm$ 8.9} & 76.2 $\pm$ 8.3 \\
Synth IntraR 1.0 & 74.9 $\pm$ 8.3 & 74.3 $\pm$ 7.6 & \textbf{77.5 $\pm$ 7.0} & 76.0 $\pm$ 6.7 & 75.0 $\pm$ 7.9 & 76.9 $\pm$ 6.1 \\
\specialrule{0.15pt}{1pt}{0.4pt}
\specialrule{0.15pt}{0.4pt}{1pt}
Real-World Avg. & 79.7 $\pm$ 5.0 & 79.6 $\pm$ 5.4 & \textbf{80.2 $\pm$ 5.0} & 79.7 $\pm$ 5.3 & 79.9 $\pm$ 4.7 & 79.9 $\pm$ 5.4 \\
Synth (No Shift) Avg. & 93.9 $\pm$ 0.2 & 93.9 $\pm$ 0.1 & \textbf{94.0 $\pm$ 0.2} & 93.9 $\pm$ 0.2 & 93.9 $\pm$ 0.2 & 93.9 $\pm$ 0.2 \\
Synth Avg. & 71.8 $\pm$ 11.0 & 72.2 $\pm$ 10.7 & 71.5 $\pm$ 10.7 & \textbf{75.0 $\pm$ 9.7} & 71.3 $\pm$ 10.3 & \textbf{75.0 $\pm$ 9.4} \\
\bottomrule
\end{tabular}
}
\end{scriptsize}
\end{table*}

We also evaluate the real-world efficacy of the CALDA framework using six real-world multi-variate time series datasets. These include four human activity recognition datasets (UCI HAR \cite{anguita2013public},
UCI HHAR \cite{stisen2015smartdevices},
WISDM AR \cite{kwapisz2011wisdmar},
and WISDM AT \cite{lockhart2011wisdmat}) and two multivariate EMG datasets collected using a Myo armband (Myo EMG \cite{cote2019emgmyo} and NinaPro Myo \cite{pizzolato2017ninaprodb5}). The Supplemental Material contain details about dataset pre-processing and hyperparameter tuning.

\subsection{Ablation Studies}
Using a set of ablation studies, we identify the appropriate instantiations of our CALDA framework to compare with the baseline methods. We compare instantiations across each component of our framework: (1) including the pseudo-labeled target domain (P), (2) using within-source (CALDA-In), any-source (CALDA-Any), or cross-source (CALDA-XS) examples for each query, and (3) random sampling (R) versus hard sampling (H). A one-sided paired student's t-test indicates whether accuracy improvements are statistically significant.

\subsubsection{Pseudo-Labeled Target Domain}
To determine whether to include pseudo-labeled target domain data or not, we compare experimental results with and without pseudo labeling, respectively. Comparing Tables~\ref{table:ablation_pseudo} and \ref{table:ablation_selection}, we find that regardless of the choices for the other components in our framework, pseudo labeling generally performs worse than without pseudo-labeling on the real-world datasets, i.e., the corresponding values in Table~\ref{table:ablation_pseudo} are significantly lower than those in Table~\ref{table:ablation_selection} ($p<0.01$). Interestingly, random sampling typically performs better than hard sampling when using pseudo labels. This is likely because incorrectly pseudo-labeled target data may often be selected during hard sampling and thereby degrade contrastive learning performance. Random sampling helps to partially reduce this performance degradation by reducing the likelihood that the pseudo-labeled target domain is used in contrastive learning. However, we obtain even better performance by explicitly excluding the target domain in contrastive learning, as shown in Table~\ref{table:ablation_selection}. Thus, in subsequent experiments, we exclude pseudo-labeling. Future work is required to determine which pseudo labeling techniques perform well in a time series context. For the synthetic datasets, we do not observe significant differences between pseudo labeling and not pseudo labeling (see Supplemental Material).

\subsubsection{Multiple Source Domains}
We next analyze the results in Table~\ref{table:ablation_selection} to determine whether to use within-source, any-source, or cross-source examples. Starting with the results on real-world datasets, the best method is consistently among the CALDA-Any and CALDA-XS variants. In particular, CALDA-Any,R is the best-performing instantiation on average with the two CALDA-XS variants ranking second, though no variation offers statistically significant improvement. Thus, we construct and include results on the additional synthetic datasets. As expected, all methods perform almost identically for no domain shift. However, when we include various types of synthetic domain shifts, we see significant differences. CALDA-Any,H and CALDA-XS,H are the best methods on average and are both significantly better than CALDA-In,R and CALDA-In,H ($p<0.01$). Thus, a similar trend emerges from both the real-world data and synthetic data: CALDA-Any and CALDA-XS instantiations outperform CALDA-In. Since CALDA-In ignores cross-source information by only utilizing within-source examples, the other instantiations performing better validates our hypothesis that leveraging cross-source information can yield improved transfer. Thus, we conclude that CALDA-Any and CALDA-XS, which both leverage cross-source information, are the two most promising methods of selecting source domain examples for our framework.

\subsubsection{Pair Selection by Difficulty}
The choice between selecting examples by hard or random sampling is more variable than the above design decisions. Comparing hard sampling with random sampling on the real datasets in Table~\ref{table:ablation_selection}, we observe random sampling for CALDA-Any to perform significantly better than hard sampling ($p<0.05)$. However, random vs. hard sampling differences for the other methods are not statistically significant. On the synthetic datasets with domain shifts, we observe the opposite: hard sampling versions of CALDA-Any and CALDA-XS are significantly better than random sampling ($p<0.01$). This may indicate this design choice depends on the type of data and domain shift.

\begin{figure}
\centering
\includegraphics[width=0.7\linewidth]{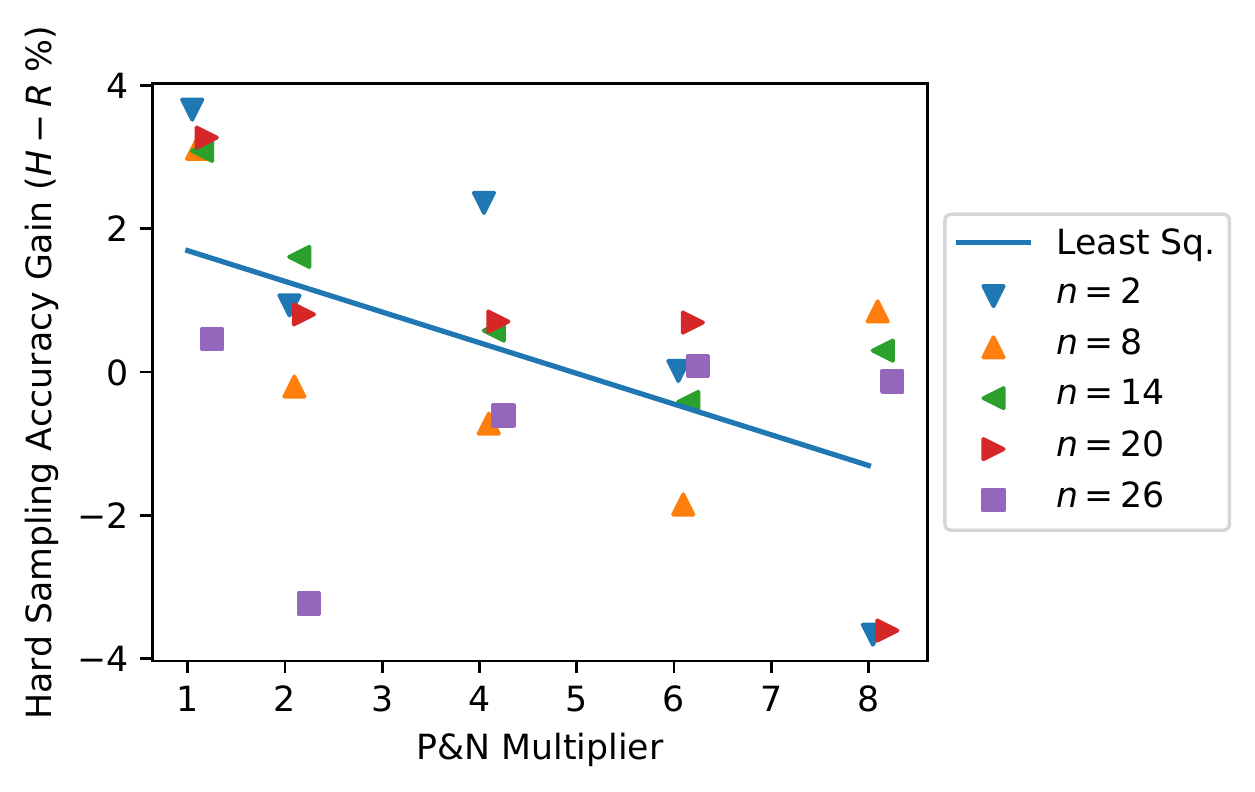}
\caption[Comparing hard vs. random sampling when increasing the number of positives and negatives]{Comparing hard vs. random sampling as $|P|$ and $|N|$ increase.}
\label{fig:ablation_sampling}
\end{figure}

We further investigate hard vs. random sampling by running an additional set of experiments for CALDA-XS on WISDM AR, the dataset where CALDA-XS,H performed better than all other instantiations, and specifically, where it outperformed CALDA-XS,R. The results as we increase the number of positives and negatives (two of the hyperparameters) via a positive/negative multiplier are shown in Figure~\ref{fig:ablation_sampling}. On the left, we observe CALDA-XS,H outperforms CALDA-XS,R. Moving to the right, the performance gain from hard sampling reduces, which is expected since hard and random sampling are no different in the limit of using all positive and negative examples in each mini-batch. We conclude that hard sampling may yield an improvement over random sampling in some situations, particularly those for which the optimal hyperparameters for a dataset include relatively few positives and negatives (such as is the case on the WISDM AR dataset).

Finally, because CALDA-Any,R performed best on the real-world data and CALDA-XS,H was tied for second best on the real-world data and tied for best on the synthetic datasets, we select these two methods as the most-promising instantiations of our framework for subsequent experiments.

\subsection{MS-UDA: CALDA vs. Baselines}

To measure the CALDA's performance improvement, we compare the two most promising instantiations, CALDA-Any,R and CALDA-XS,H, with baselines and prior work. We include an approximate domain adaptation lower bound that performs no adaptation during training ({\em No Adaptation}). This allows us to see how much improvement results from utilizing domain adaptation. We include an approximate domain adaptation upper bound showing the performance achievable if we did have labeled target domain data available ({\em Train on Target}). For a contrastive domain adaptation baseline, we include the Contrastive Adaptation Network ({\em CAN}) \cite{kang2019contrastive} modified to employ a time-series compatible feature extractor. Finally, we include {\em CoDATS} \cite{wilson2020codats} to see if our CALDA framework improves over prior multi-source time series domain adaptation work. The results are presented in Table~\ref{table:results}.

\begin{table*}
\centering
\caption[Comparing target domain accuracy of the most-promising CALDA instantiations with baselines]{Comparing target domain accuracy of the most-promising CALDA instantiations with baselines. Bold denotes CALDA outperforming baselines. Underline denotes highest accuracy in each row.} %
\label{table:results}
\begin{scriptsize}
{\renewcommand{\arraystretch}{1.4}
\begin{tabular}{ccccccc}
\toprule
Dataset & No Adaptation & CAN & CoDATS & \textit{CALDA-Any,R} & \textit{CALDA-XS,H} & Train on Target \\
\midrule
UCI HAR & 88.8 $\pm$ 4.2 & 89.0 $\pm$ 3.7 & 92.8 $\pm$ 3.3 & \textbf{93.1 $\pm$ 2.3} & \textbf{\underline{93.4 $\pm$ 2.5}} & 99.6 $\pm$ 0.1 \\
UCI HHAR & 76.9 $\pm$ 6.3 & 77.5 $\pm$ 5.3 & 88.6 $\pm$ 3.9 & \textbf{\underline{89.8 $\pm$ 3.7}} & \textbf{89.4 $\pm$ 4.0} & 98.9 $\pm$ 0.2 \\
WISDM AR & 72.1 $\pm$ 8.0 & 61.3 $\pm$ 7.5 & 78.0 $\pm$ 8.4 & \textbf{80.2 $\pm$ 7.1} & \textbf{\underline{81.4 $\pm$ 7.9}} & 96.5 $\pm$ 0.1 \\
WISDM AT & 69.9 $\pm$ 7.1 & 66.2 $\pm$ 9.6 & 69.7 $\pm$ 6.6 & \textbf{\underline{72.1 $\pm$ 7.5}} & \textbf{71.0 $\pm$ 8.6} & 98.8 $\pm$ 0.1 \\
Myo EMG & 77.4 $\pm$ 5.2 & 74.1 $\pm$ 6.3 & 82.2 $\pm$ 5.4 & \textbf{\underline{83.8 $\pm$ 5.6}} & \textbf{83.3 $\pm$ 5.3} & 97.7 $\pm$ 0.1 \\
NinaPro Myo & 54.8 $\pm$ 3.6 & 56.8 $\pm$ 3.2 & 55.9 $\pm$ 5.0 & \textbf{\underline{57.7 $\pm$ 3.8}} & 56.0 $\pm$ 3.9 & 77.8 $\pm$ 1.3 \\
\hline
Synth SW & 62.2 $\pm$ 11.0 & \underline{78.2 $\pm$ 7.7} & 63.8 $\pm$ 10.6 & 71.5 $\pm$ 10.7 & 75.0 $\pm$ 9.5 & 93.7 $\pm$ 0.2 \\
Synth 1GMM & 88.5 $\pm$ 3.4 & 90.1 $\pm$ 4.2 & 91.4 $\pm$ 2.8 & \textbf{\underline{92.3 $\pm$ 3.7}} & 91.4 $\pm$ 3.4 & 98.4 $\pm$ 0.5 \\
Synth 2GMM & 86.3 $\pm$ 6.5 & 89.4 $\pm$ 5.7 & 91.8 $\pm$ 7.8 & \textbf{93.6 $\pm$ 5.1} & \textbf{\underline{94.0 $\pm$ 4.7}} & 100.0 $\pm$ 0.0 \\
Synth 3GMM & 80.1 $\pm$ 9.0 & 83.8 $\pm$ 9.1 & 91.3 $\pm$ 8.1 & \textbf{91.4 $\pm$ 6.9} & \textbf{\underline{92.2 $\pm$ 7.2}} & 100.0 $\pm$ 0.0 \\
\specialrule{0.15pt}{1pt}{0.4pt}
\specialrule{0.15pt}{0.4pt}{1pt}
Real-World Avg. & 73.9 $\pm$ 5.8 & 71.3 $\pm$ 6.0 & 78.6 $\pm$ 5.4 & \textbf{\underline{80.2 $\pm$ 5.0}} & \textbf{79.9 $\pm$ 5.4} & 94.9 $\pm$ 0.3 \\
Synth Avg. & 79.3 $\pm$ 7.5 & 85.4 $\pm$ 6.7 & 84.6 $\pm$ 7.3 & \textbf{87.2 $\pm$ 6.6} & \textbf{\underline{88.2 $\pm$ 6.2}} & 98.0 $\pm$ 0.2 \\
\bottomrule
\end{tabular}
}
\end{scriptsize}
\end{table*}

We first examine the No Adaptation and Train on Target baselines, which train only on the source domain data or train directly on the target domain data respectively. The performance of the No Adaptation baseline can be viewed as an approximate measure of domain adaptation difficulty, where lower No Adaptation performance indicates a more challenging problem, i.e., with a larger domain gap between the source domains and the target domain. Accordingly, we can identify WISDM AR and WISDM AT as the most challenging activity recognition datasets and NinaPro Myo as the most challenging EMG dataset. In contrast, Train on Target performs well on all but one dataset. It does this well by ``cheating'', i.e., looking at the target domain labels and thereby eliminating the domain gap. However, the NinaPro Myo dataset is challenging enough that even with no domain gap, we cannot obtain near-perfect accuracy.

Next, we compare CALDA-Any,R and CALDA-XS,H with the No Adaptation and CoDATS baselines. One of the two CALDA instantiations always performs best. Similarly, CALDA-Any,R and CALDA-XS,H significantly outperform both No Adaptation and CoDATS across all datasets ($p<0.01$). On the real-world datasets, the largest improvement over CoDATS is 2.4\% on WISDM AT for CALDA-Any,R and 3.4\% on WISDM AR for CALDA-XS,H. The largest improvement over No Adaptation is 12.9\% and 12.5\%, respectively, on UCI HHAR. On average, we observe a 1.6\% and 1.3\% improvement of these two CALDA instantiations over CoDATS and a 6.3\% and 6.0\% improvement over No Adaptation, respectively. On the synthetic datasets, these improvements are even larger: 7.7\% and 11.2\% improvement over CoDATS and 9.3\% and 12.8\% improvement over no Adaptation. These experimental results across a variety of real world and synthetic time-series datasets confirm the benefit of utilizing cross-source information through the CALDA framework for time-series multi-source domain adaptation.

Finally, we compare CALDA-Any,R and CALDA-XS,H with the contrastive domain adaptation baseline CAN. Both CALDA instantiations significantly outperform CAN on all of the real-world time series datasets ($p<0.01$). The largest improvements over CAN on the real-world datasets are 18.9\% for CALDA-Any,R and 20.1\% for CALDA-XS,H on WISDM AR. On average, we observe an 8.9\% and 8.6\% improvement in the two CALDA instantiations over CAN.
On the synthetic datasets, CAN yields the strongest results on the simple synthetic data (SW). CAN relies on clustering for pseudo-labeling target domain data, which works well for the simple, synthetically-generated Gaussian domain shifts.
CALDA outperforms the other methods for the more complex GMM scenarios ($p<0.05$), and the amount of improvement grows with the data complexity.
These results indicate that while CAN may be successful with some types of domain shifts such as those found in image datasets or clustered synthetic time series, we find that CALDA better handles the domain shifts found in complex real-world time series datasets.

\subsection{MS-UDA with Weak Supervision}

We additionally study whether our framework yields improved results for domain adaptation with weak supervision. First, we simulate obtaining target domain label proportions by estimating these proportions on the target domain training set and incorporate our weak supervision regularizer into each method to leverage this additional information. Following this, we determine the sensitivity of each method to noise in the estimated label proportions since, for example, if these label proportions are acquired from participants' self-reports, there will be some error in the reported proportions.

\subsubsection{CALDA with Weak Supervision}
We compare the two most promising instantiations of CALDA-WS (\textit{CALDA-XS,H,WS} and \textit{CALDA-Any,R,WS}) with \textit{CoDATS-WS}. The results are shown in Table~\ref{table:results_ws_combined}. Similar to no weak supervision, CALDA-Any,R,WS improves over both No Adaptation and CoDATS-WS across all datasets and CALDA-XS,H,WS in all except one case ($p<0.01$). On average we observe a 4.1\% and 3.6\% improvement of the two CALDA instantiations over CoDATS-WS and a 9.8\% and 9.3\% improvement over No Adaptation on the real-world datasets. On the synthetic datasets, we observe a 12.3\% and 16.5\% improvement of CALDA over both CoDATS-WS and No Adaptation. These results demonstrate the efficacy of CALDA for the domain adaptation when incorporating weak supervision.

By comparing Table~\ref{table:results} with \ref{table:results_ws_combined}, we can measure the benefit of weak supervision. In all cases, the CALDA instantiations with weak supervision significantly improve over CALDA without weak supervision ($p<0.01$). On the real-world datasets, this is also the case with CoDATS: CoDATS-WS improves over CoDATS ($p<0.05$). On the real-world datasets, we observe a 3.5\% and 3.3\% improvement for the CALDA instantiations by including weak supervision. On the synthetic datasets, these differences are 3.0\% and 3.7\%. We observe the largest performance gains from utilizing weak supervision on the two unbalanced datasets: 5.0\% and 3.3\% improvements of the two instantiations on WISDM AR and 10.8\% and 12.3\% improvements on WISDM AT. Because these datasets are unbalanced, larger differences on these datasets are expected since our weak supervision regularization term capitalizes on label distribution differences among the domains. These gains demonstrate: (1) the benefit of leveraging weak supervision for domain adaptation when available, and (2) the observation that CALDA yields improvements over prior work,  even for this related problem setting.

\begin{table*}
\centering
\caption[Comparing target domain accuracy for domain adaptation methods utilizing weak supervision]{Comparing target domain accuracy for domain adaptation methods utilizing weak supervision. Bold denotes CALDA outperforming baselines. Underline denotes highest accuracy in each row.}
\label{table:results_ws_combined}
\begin{scriptsize}
{\renewcommand{\arraystretch}{1.4}
\begin{tabular}{cccccc}
\toprule
Dataset & No Adaptation & CoDATS-WS & \textit{CALDA-Any,R,WS} & \textit{CALDA-XS,H,WS} & Train on Target \\
\midrule
UCI HAR & 88.8 $\pm$ 4.2 & 92.9 $\pm$ 3.2 & \textbf{94.8 $\pm$ 1.8} & \textbf{\underline{95.5 $\pm$ 2.1}} & 99.6 $\pm$ 0.1 \\
UCI HHAR & 76.9 $\pm$ 6.3 & 88.2 $\pm$ 4.6 & \textbf{\underline{90.2 $\pm$ 3.8}} & \textbf{89.8 $\pm$ 4.1} & 98.9 $\pm$ 0.2 \\
WISDM AR & 72.1 $\pm$ 8.0 & 84.9 $\pm$ 7.2 & \textbf{\underline{85.2 $\pm$ 6.9}} & 84.7 $\pm$ 7.0 & 96.5 $\pm$ 0.1 \\
WISDM AT & 69.9 $\pm$ 7.1 & 72.1 $\pm$ 10.3 & \textbf{82.9 $\pm$ 7.3} & \textbf{\underline{83.3 $\pm$ 7.1}} & 98.8 $\pm$ 0.1 \\
Myo EMG & 77.4 $\pm$ 5.2 & 79.5 $\pm$ 5.7 & \textbf{\underline{85.1 $\pm$ 4.6}} & \textbf{84.7 $\pm$ 4.6} & 97.7 $\pm$ 0.1 \\
NinaPro Myo & 54.8 $\pm$ 3.6 & 54.9 $\pm$ 4.2 & \textbf{\underline{58.8 $\pm$ 3.8}} & \textbf{56.0 $\pm$ 4.5} & 77.8 $\pm$ 1.3 \\
\hline
Synth SW & 62.2 $\pm$ 11.0 & 62.2 $\pm$ 11.6 & \textbf{74.5 $\pm$ 11.3} & \textbf{\underline{78.8 $\pm$ 9.2}} & 93.7 $\pm$ 0.1 \\
Synth 1GMM & 88.5 $\pm$ 3.4 & 90.4 $\pm$ 3.0 & \textbf{\underline{91.7 $\pm$ 2.5}} & \textbf{90.8 $\pm$ 3.7} & 98.4 $\pm$ 0.5 \\
Synth 2GMM & 86.3 $\pm$ 6.5 & 86.4 $\pm$ 12.9 & \textbf{92.3 $\pm$ 7.3} & \textbf{\underline{93.7 $\pm$ 7.2}} & 100.0 $\pm$ 0.0 \\
Synth 3GMM & 80.1 $\pm$ 9.0 & 87.7 $\pm$ 9.2 & 84.1 $\pm$ 13.5 & \textbf{\underline{89.1 $\pm$ 10.9}} & 100.0 $\pm$ 0.0 \\
\specialrule{0.15pt}{1pt}{0.4pt}
\specialrule{0.15pt}{0.4pt}{1pt}
Real-World Avg. & 73.9 $\pm$ 5.8 & 79.6 $\pm$ 5.9 & \textbf{\underline{83.7 $\pm$ 4.7}} & \textbf{83.2 $\pm$ 4.9} & 94.9 $\pm$ 0.3 \\
Synth Avg. & 62.2 $\pm$ 11.0 & 62.2 $\pm$ 11.6 & \textbf{85.7 $\pm$ 8.7} & \textbf{\underline{88.1 $\pm$ 7.8}} & 93.7 $\pm$ 0.2 \\
\bottomrule
\end{tabular}

}
\end{scriptsize}
\end{table*}

\subsubsection{Sensitivity of Weak Supervision to Noise}

By leveraging weak supervision, we were able to improve performance. However, in the above experiments, we simulated obtaining target domain label proportions by estimating those proportions exactly on the target domain training dataset. Now we perform additional experiments to determine how robust these methods are to noise in the estimated label proportions. Since weak supervision has the greatest effect when label distributions differ among domains, we compare these methods with various noise budgets on the unbalanced WISDM datasets. A noise budget of 0.1 indicates that approximately 10\% of the class labels can be redistributed. In the case of human activity recognition, if all hours of the day correspond with an activity, then this represents 10\% of the day being attributed to an incorrect activity when self-reporting label proportions for weak supervision. The results are shown in Table~\ref{table:ws_noise}. Note that label proportions are redistributed according to the noise budget and then re-normalized so the proportions remain a valid distribution. In the table we provide the \textit{True Post-Norm. Noise} column to validate that the true post-normalized noise on average is close to the desired noise budget.

CALDA typically outperforms CoDATS both with and without weak supervision on the WISDM AR dataset (the final row corresponds to methods without weak supervision). Similarly, the best method in each row is always one of the two CALDA instantiations. We observe that even with a noise budget of 0.1, CALDA-WS and CoDATS-WS perform better than CALDA and CoDATS without weak supervision. However, beyond this threshold, we find additional noise degrades performance on WISDM AR. From these results, we conclude the maximum acceptable noise level for weak supervision on WISDM AR is between 0.1 and 0.2. In the case of WISDM AT, the two CALDA instantiations outperform CoDATS both without weak supervision and with weak supervision regardless of the amount of noise. We find that it takes a noise budget of 0.1 before CoDATS-WS degrades to the performance of CoDATS without weak supervision. However, for CALDA-WS, the noise budget can be as large as 0.4 before it degrades to the performance of CALDA without weak supervision.

Overall, on these two datasets, we find that leveraging both weak supervision and cross-source label information can yield improved domain adaptation performance, even with some noise in the weak supervision label information. Though, the acceptable amount of noise depends on the dataset. On both datasets, CoDATS-WS requires a noise level of no more than approximately 0.1, and both CALDA-WS instantiations have similar limits on WISDM AR. However, on WISDM AT, the noise budget for either CALDA-WS instantiation can be as high as 0.4 -- four times that of CoDATS-WS. Thus, we conclude that our CALDA framework improves over CoDATS with and without weak supervision and also that our CALDA framework can yield higher robustness to noise in the weak supervision label information on some datasets.

\begin{table*}
\centering
\caption[Weak supervision sensitivity to noise]{Weak supervision sensitivity to noise. Bold denotes higher accuracy than CoDATS. Underlining denotes best method in each row.}
\label{table:ws_noise}
\begin{scriptsize}
{\renewcommand{\arraystretch}{1.4}
\begin{tabular}{ccccccc}
\toprule
Dataset & Weak Supervision & Noise Budget & True Post-Norm. Noise & CoDATS & \textit{CALDA-XS,H} & \textit{CALDA-Any,R} \\
\midrule
\multirow{5}{*}{WISDM AR} & \multirow{5}{*}{Yes} & 0.0 & 0.0 & 84.9 $\pm$ 7.2 & 84.7 $\pm$ 7.0 & \underline{\textbf{85.2 $\pm$ 6.9}} \\
& & 0.05 & 0.06 & 82.9 $\pm$ 7.4 & \textbf{83.4 $\pm$ 6.2} & \underline{\textbf{84.0 $\pm$ 6.0}} \\
& & 0.1 & 0.12 & 79.1 $\pm$ 8.4 & \underline{\textbf{83.1 $\pm$ 7.7}} & \textbf{81.8 $\pm$ 6.8} \\
& & 0.2 & 0.22 & 74.6 $\pm$ 8.8 & \underline{\textbf{78.9 $\pm$ 6.7}} & \textbf{78.4 $\pm$ 8.1} \\
& & 0.4 & 0.38 & 64.9 $\pm$ 9.7 & \textbf{68.9 $\pm$ 8.8} & \underline{\textbf{69.9 $\pm$ 9.2}} \\
\hline
WISDM AR & No & N/A & N/A & 78.0 $\pm$ 8.4 & \underline{\textbf{81.4 $\pm$ 7.9}} & \textbf{80.2 $\pm$ 7.1} \\
\specialrule{0.15pt}{1pt}{0.4pt}
\specialrule{0.15pt}{0.4pt}{1pt}
\multirow{5}{*}{WISDM AT} & \multirow{5}{*}{Yes} & 0.0 & 0.0 & 72.1 $\pm$ 10.3 & \underline{\textbf{83.3 $\pm$ 7.1}} & \textbf{82.9 $\pm$ 7.3} \\
& & 0.05 & 0.07 & 71.0 $\pm$ 11.9 & \textbf{81.8 $\pm$ 6.6} & \underline{\textbf{82.4 $\pm$ 7.3}} \\
& & 0.1 & 0.13 & 69.9 $\pm$ 14.3 & \textbf{81.1 $\pm$ 7.9} & \underline{\textbf{82.7 $\pm$ 7.4}} \\
& & 0.2 & 0.23 & 65.6 $\pm$ 11.4 & \textbf{78.3 $\pm$ 8.3} & \underline{\textbf{79.2 $\pm$ 7.8}} \\
& & 0.4 & 0.40 & 56.3 $\pm$ 12.8 & \underline{\textbf{72.0 $\pm$ 8.6}} & \textbf{71.3 $\pm$ 8.3} \\
\hline
WISDM AT & No & N/A & N/A & 69.7 $\pm$ 6.6 & \textbf{71.0 $\pm$ 8.6} & \underline{\textbf{72.1 $\pm$ 7.5}} \\
\bottomrule
\end{tabular}
}
\end{scriptsize}
\end{table*}

\subsection{Validating Assumptions}

Now we examine the validity of two key assumptions.

\subsubsection{Importance of the Adversary}

Using the CALDA framework, we investigated various design choices for how to use contrastive learning for domain adaptation. However, we made the assumption that adversarial learning is an important component for each of these instantiations. Here we illustrate why. For the two most-promising instantiations, we run experiments when excluding the adversary. The results on the real-world datasets are shown in Table~\ref{table:ablation_noadv}. The methods with an adversary are far superior to those when we exclude the adversary ($p<0.01$). This justifies our inclusion of the adversary.

\begin{table*}
\centering
\caption[Ablation study comparing CALDA with or without an adversary]{Ablation study comparing CALDA with or without an adversary. Bold denotes highest accuracy in each row.}
\label{table:ablation_noadv}
\begin{scriptsize}
{\renewcommand{\arraystretch}{1.4}
\begin{tabular}{ccccc}
\toprule
Dataset & \textit{CALDA-Any,R,NoAdv} & \textit{CALDA-XS,H,NoAdv} & \textit{CALDA-Any,R} & \textit{CALDA-XS,H} \\
\midrule
UCI HAR & 89.9 $\pm$ 3.4 & 89.8 $\pm$ 3.5 & 93.1 $\pm$ 2.3 & \textbf{93.4 $\pm$ 2.5} \\
UCI HHAR & 74.0 $\pm$ 7.2 & 74.7 $\pm$ 6.8 & \textbf{89.8 $\pm$ 3.7} & 89.4 $\pm$ 4.0 \\
WISDM AR & 70.9 $\pm$ 6.8 & 72.1 $\pm$ 6.2 & 80.2 $\pm$ 7.1 & \textbf{81.4 $\pm$ 7.9} \\
WISDM AT & 70.9 $\pm$ 7.7 & 70.2 $\pm$ 7.9 & \textbf{72.1 $\pm$ 7.5} & 71.0 $\pm$ 8.6 \\
Myo EMG & 76.2 $\pm$ 5.2 & 76.2 $\pm$ 5.0 & \textbf{83.8 $\pm$ 5.6} & 83.3 $\pm$ 5.3 \\
NinaPro Myo & 53.4 $\pm$ 4.0 & 51.4 $\pm$ 4.4 & \textbf{57.7 $\pm$ 3.8} & 56.0 $\pm$ 3.9 \\
\hline
Average & 73.2 $\pm$ 5.8 & 73.1 $\pm$ 5.7 & \textbf{80.2 $\pm$ 5.0} & 79.9 $\pm$ 5.4 \\
\bottomrule
\end{tabular}

}
\end{scriptsize}
\end{table*}

\begin{table*}
\centering
\caption[Comparing domain adaptation performance excluding unlabeled target domain data during training]{Comparing domain adaptation performance excluding unlabeled target domain data during training. Bold denotes methods outperforming CoDATS-DG and No Adaptation baselines. Underline denotes highest accuracy.}
\label{table:results_dg}
\begin{scriptsize}
{\renewcommand{\arraystretch}{1.4}
\begin{tabular}{cccccccc}
\toprule
Dataset & No Adaptation & CoDATS-DG & Sleep-DG & AFLAC-DG & CALDG-Any,R & CALDG-XS,H & Train on Target \\
\midrule
UCI HAR & 88.8 $\pm$ 4.2 & 88.4 $\pm$ 3.7 & 87.0 $\pm$ 4.8 & \textbf{89.3 $\pm$ 4.6} & \textbf{89.5 $\pm$ 3.8} & \textbf{\underline{90.0 $\pm$ 3.7}} & 99.6 $\pm$ 0.1 \\
UCI HHAR & \underline{76.9 $\pm$ 6.3} & 76.0 $\pm$ 6.1 & 75.4 $\pm$ 6.7 & 76.6 $\pm$ 6.4 & 76.6 $\pm$ 6.4 & 76.2 $\pm$ 6.9 & 98.9 $\pm$ 0.2 \\
WISDM AR & \underline{72.1 $\pm$ 8.0} & 66.9 $\pm$ 8.7 & 66.9 $\pm$ 8.9 & 70.9 $\pm$ 7.8 & 68.9 $\pm$ 8.8 & 70.2 $\pm$ 7.8 & 96.5 $\pm$ 0.1 \\
WISDM AT & 69.9 $\pm$ 7.1 & 69.7 $\pm$ 7.8 & 68.3 $\pm$ 8.9 & 69.7 $\pm$ 6.6 & \textbf{\underline{70.7 $\pm$ 7.4}} & \textbf{\underline{70.7 $\pm$ 7.7}} & 98.8 $\pm$ 0.1 \\
Myo EMG & 77.4 $\pm$ 5.2 & 73.0 $\pm$ 5.3 & 73.9 $\pm$ 5.9 & 74.3 $\pm$ 5.5 & \textbf{\underline{78.4 $\pm$ 4.8}} & 76.8 $\pm$ 5.8 & 97.7 $\pm$ 0.1 \\
NinaPro Myo & 54.8 $\pm$ 3.6 & 50.6 $\pm$ 4.3 & 50.4 $\pm$ 4.7 & 51.1 $\pm$ 3.5 & \textbf{\underline{55.1 $\pm$ 3.7}} & 49.8 $\pm$ 4.6 & 77.8 $\pm$ 1.3 \\
\hline
Synth InterT 10 & 62.6 $\pm$ 18.8 & 68.4 $\pm$ 13.9 & \textbf{68.7 $\pm$ 13.9} & 67.6 $\pm$ 13.4 & \textbf{69.8 $\pm$ 16.6} & \textbf{\underline{71.2 $\pm$ 15.9}} & 93.4 $\pm$ 0.2 \\
Synth InterR 1.0 & 52.4 $\pm$ 7.8 & 53.3 $\pm$ 8.0 & \textbf{53.7 $\pm$ 7.0} & \textbf{66.6 $\pm$ 7.7} & \textbf{65.4 $\pm$ 10.5} & \textbf{\underline{75.5 $\pm$ 8.1}} & 94.0 $\pm$ 0.0 \\
Synth IntraT 10 & 70.6 $\pm$ 8.5 & 66.5 $\pm$ 10.8 & 68.2 $\pm$ 10.2 & 68.5 $\pm$ 7.5 & \textbf{73.2 $\pm$ 8.7} & \textbf{\underline{74.5 $\pm$ 8.7}} & 93.7 $\pm$ 0.2 \\
Synth IntraR 1.0 & 63.3 $\pm$ 9.0 & 59.9 $\pm$ 7.1 & 62.4 $\pm$ 6.3 & \textbf{64.0 $\pm$ 7.8} & \textbf{\underline{77.3 $\pm$ 8.0}} & \textbf{76.6 $\pm$ 6.1} & 93.6 $\pm$ 0.2 \\
\specialrule{0.15pt}{1pt}{0.4pt}
\specialrule{0.15pt}{0.4pt}{1pt}
Real-World Avg. & \underline{73.9 $\pm$ 5.8} & 71.4 $\pm$ 6.1 & 71.0 $\pm$ 6.7 & 72.7 $\pm$ 5.8 & 73.8 $\pm$ 5.9 & 73.1 $\pm$ 6.1 & 94.9 $\pm$ 0.3 \\
Synth Avg. & 62.2 $\pm$ 11.0 & 62.0 $\pm$ 9.9 & \textbf{63.3 $\pm$ 9.3} & \textbf{66.7 $\pm$ 9.1} & \textbf{71.4 $\pm$ 11.0} & \textbf{\underline{74.5 $\pm$ 9.7}} & 93.7 $\pm$ 0.2 \\
\bottomrule
\end{tabular}

}
\end{scriptsize}
\end{table*}

\subsubsection{Importance of Unlabeled Target Domain Data}

Finally, in the problem of unsupervised domain adaptation, we have unlabeled target domain data available for use during training. Unsupervised domain adaptation methods typically make the assumption that these data are useful for improving target-domain performance. Here, both CoDATS and CALDA leverage this data via adversarial learning, which as observed above is vital to domain adaptation performance. However, another alternative is to only perform adversarial learning among the multiple source domains and exclude the target-domain unlabeled data, i.e., promote domain-invariant features among only the multiple source domains through the adversarial loss. This is related to the problem of domain generalization \cite{blanchard2011dg}. The results for the corresponding \textit{CoDATS-DG}, \textit{CALDG-XS,H}, and \textit{CALDG-Any,R} methods are shown in Table~\ref{table:results_dg}. For comparison, we also include two domain generalization methods \textit{Sleep-DG} \cite{zhao2017icml} and \textit{AFLAC-DG} \cite{akuzawa2019adversarial}. Comparing Tables~\ref{table:results} and \ref{table:results_dg}, on the real-world datasets we observe including the unlabeled data yields significantly higher accuracy of CoDATS, CALDA-Any,R, and CALDA-XS,H ($p<0.01$). This is similarly true on the synthetic data, but the differences are not large enough to be significant. However, on both real and synthetic datasets, CALDG-Any,R and CALDG-XS,H are significantly better than CoDATS-DG, Sleep-DG, and AFLAC-DG ($p<0.01$). On the synthetic datasets, they are similarly better than No Adaptation ($p<0.01$). In contrast, on the real-world datasets, No Adaptation performs the best on average, though not significantly different than CALDG-Any,R. From these experiments, we conclude that the unlabeled target domain data makes a significant contribution to our proposed CALDA method. In addition, contrastive learning appears to benefit the problem of domain generalization as well as domain adaptation, though we leave a more detailed investigation to future work.

\section{Conclusions and Future Work}

We propose a novel time series MS-UDA framework CALDA, drawing on the principles of both adversarial and contrastive learning. This approach seeks to improve transfer by leveraging cross-source information, which is ignored by prior work time series work. We investigate design decisions for incorporating contrastive learning into multi-source domain adaptation, including how to select examples from multiple domains, whether to include the target domain, and whether to utilize example difficulty. We observe that CALDA improves performance over prior work on a variety of real-world and synthetic time-series datasets both with and without weak supervision. In the weak supervision case, we additionally find the method is robust to label proportion noise. We also validated that both the adversary and unlabeled target domain data yield significant contribution to domain adaptation performance.

We observe the influence of hyperparameters on CALDA's performance. In a UDA setting, we do not rely on labeled source data to guide hyperparameter selection. Instead, methods such as those proposed by Saito et al. \cite{Saito21} and You et al. \cite{You19} can be leveraged to fine-tune CALDA's learned features toward a source density that maximizes discriminability.

In Section~\ref{section:MSUDA}, we define time series as containing values that appear at uniform time intervals. In situations where time series data are unevenly spaced, such as electronic health record analysis or harmonizing multi-source data, data can be preprocessed to achieve uniformity \cite{Bharambe2021}. Alternatively, the model can reason about the non-uniform time delays by incorporating them into the model as a separate parameter \cite{Braun2022,Koss2022}.

Additional future work includes
examining the use of CALDA for cross-modality adaptation \cite{Deldari2022}.
We will also develop data augmentation compatible with time series domain adaptation and pseudo labeling techniques viable for the large domain gaps observed in time series to see if these yield further improvements in transfer.

\ifCLASSOPTIONcompsoc
  \section*{Acknowledgments}
\else
  \section*{Acknowledgment}
\fi

This material is based upon work supported by the National Science Foundation under Grant No. 1543656 and by the National Institutes of Health under Grant No. R01EB009675. %

\ifCLASSOPTIONcaptionsoff
  \newpage
\fi


\bibliographystyle{IEEEtran}
\bibliography{IEEEabrv,bibliography_nourl}

\newpage
\begin{IEEEbiographynophoto}{Garrett Wilson}
received his BS in Engineering from Walla Walla University in 2016. He received a PhD in Computer Science from Washington State University. His research interests include transfer learning, generative adversarial networks, and time series.
\end{IEEEbiographynophoto}

\begin{IEEEbiographynophoto}{Janardhan Rao Doppa} is a Huie-Rogers Endowed Chair Associate Professor at Washington State University. He received his PhD from Oregon State University. His research interests include machine learning with a focus on small-data setting and robustness.
\end{IEEEbiographynophoto}

\begin{IEEEbiographynophoto}{Diane J. Cook}
is a Regents and Huie-Rogers Professor at Washington State University. She received her PhD from the University of Illinois. Her research interests include machine learning, pervasive computing, and design of automated strategies for health monitoring and intervention.
\end{IEEEbiographynophoto}

\enlargethispage{-3.5in}

\newpage

\appendices

\section{Supplemental Material}

\IEEEPARstart{H}{ere} we further document the dataset pre-processing, model architecture, hyperparameter tuning, and training algorithm for the CALDA framework and experiments from the main paper. We also provide a few additional tables which could not be included in the main paper due to space constraints, further corroborating the conclusions given in the main paper.

\section{Experimental Setup}

\subsection{Hyperparameter Tuning}
For CALDA, CoDATS, and No Adaptation, we performed hyperparameter tuning with a random search over the following space: learning rate $lr \in \{ 0.00001, 0.0001, 0.001 \}$, max number of positives selected $|P| \in \{ 5, 10 \}$, max number of negatives selected $|N| = r \cdot |P|$ with the negative to positive ratio $r \in \{ 2, 4 \}$, contrastive loss weight $\lambda_c \in \{ 1.0, 10.0, 100.0 \}$, and contrastive loss temperature $\tau \in \{ 0.01, 0.05, 0.1, 0.5 \}$. Tuning was performed on a limited number of adaptation problems from each dataset: 5 target users with 3 random sets of sources for each of two values of $n$ (the lowest and highest values of $n$ for each dataset). The best hyperparameters were selected based on highest accuracy on the validation set -- no method ever saw the true test set during tuning or model selection. For each dataset, the CALDA variants each use the same hyperparameters to verify the differences in results are due to the design choices rather than tuning.

For the CAN baseline, we follow the same procedure as for CALDA, but the set of hyperparameters differs. We perform a search over: base learning rate $\in \{ 0.0001, 0.001 \}$, source/target batch size $\in \{ 30, 60 \}$, alpha $\in \{ 0.0001, 0.0005, 0.001 \}$, beta $\in \{ 0.75, 1, 1.5, 2, 2.25, 2.5, 2.75, 3 \}$, and loss weight $\in \{ 0.1, 0.2, 0.3, 0.4, 0.5 \}$. These values are inclusive of the best parameters found in the original CAN experiments, but we extend the search space for these neural network hyperparameters since the time series feature extractor is sufficiently different from the image neural network. For CAN clustering, we retain the same hyperparameters as in the CAN paper: source/target clustering batch size of 600, clustering budget of 1000, and max loops of 50.

When examining the results of the hyperparameter search, we observe some trends. In the case of No adaptation, CoDATS, and the Train on Target upper bound, the methods achieved a similar performance for each set of hyperparameter choices, although the smallest learning rate yielded slightly improved results for these methods. In the case of CAN, the larger learning rates and lower loss weights performed best. However, other hyperparameters did not show clear trends. For CALDA, a larger learning rate and lower similarity weight performed best. Other hyperparameters had less impact on performance for CALDA.

We performed an additional analysis to determine how sensitive each of the $d$ datasets was to hyperparameter choices. In this experiment, we performed leave-one-dataset-out validation, selecting hyperparameters for $d-1$ datasets and observing performance for the held-out case. We observed stability for the leave-one-out tuning with the exception of UCI HAR, the simplest of the real-world datasets (evidenced by the strong No Adaptation performance). For this held-out dataset, CoDATS outperformed the other methods.

\subsection{Neural Network Model and Training}
We employ a neural network previously demonstrated to work well on time-series data \cite{wilson2020codats,wang2017strongbaseline}. To add support for contrastive learning, we include an additional contrastive head \cite{park2020jcl,cai2020negativeselection}, consisting of a single 128-unit fully-connected layer added to the model following the feature extractor. We apply the contrastive loss to the representations output by this additional layer.

Following the setup for CoDATS \cite{wilson2020codats}, each model was trained for 30,000 iterations with the Adam optimizer, a batch size of 128, and the $\lambda_d$ domain-adversarial learning rate schedule from Ganin et al. \cite{ganin2016jmlr}. Because weak supervision depends on a sufficient number of unlabeled target domain data to estimate the predicted label proportions in each batch, for weak supervision we divide the batch size equally between source domains (further split among source domains equally) and the target domain \cite{wilson2020codats}. Without weak supervision, an evenly-split batch division between domains was used. For the additional hard vs. random experiment, we instead used a batch of size 64 because, while the same trend holds for a batch of size 128, it is less visible.

For a contrastive domain adaptation baseline, we include the CAN method \cite{kang2019contrastive}. However, CAN was designed for use with image datasets. To make it compatible with time series datasets, we replace the image neural network with the same time series feature extractor used in CoDATS and CALDA. This facilitates comparing the CAN method with CALDA on time series data.

\subsection{Evaluation Metrics}
To evaluate each method, we follow the MS-UDA evaluation protocol from prior work \cite{wilson2020codats}. We select 5 different values of $n$ to determine how well each method works across various numbers of source domains. These numbers of source domains are as follows: $n \in \{ 2, 8, 14, 20, 26 \}$ for UCI HAR, $n \in \{ 2, 3, 4, 5, 6 \}$ for UCI HHAR, $n \in \{ 2, 8, 14, 20, 26 \}$ for WISDM AR, $n \in \{ 2, 12, 22, 32, 42 \}$ for WISDM AT, $n \in \{ 2, 10, 18, 26, 34 \}$ for Myo EMG, $n \in \{ 2, 4, 6, 8 \}$ for NinaPro Myo, and $n \in \{ 2, 4, 6, 8, 10 \}$ for Synth. However, to better observe overall trends across various numbers of source domains and due to space constraints, we average over these multiple values of $n$ for each dataset.

For each value of $n$, we select 10 random target domains and 3 random sets of $n$ source domains for each of those target domains. Finally, we compute the average classification accuracy of each method on the hold-out target domain test sets. Thus, each point in the results of the main paper is an average of approximately 150 experiments (30 for each value of $n$), though results for each individual value of $n$ are provided below in Tables~\ref{table:results_full} and \ref{table:results_full_synthetic_shift2}. The error given for each experiment is the average of the standard deviation over each set of 3 random sets of source domains, i.e., this error indicates the variation of each method to the 3 different source domain selections and also the 3 different random initializations of the networks. This is in contrast to typical variances given outside a multi-source domain adaptation context, where the variances only indicate variation over several random initializations since they do not have multiple source domain choices available. Overall, this evaluation procedure allows us to compare each method across a wide array of MS-UDA adaptation problems.

\subsection{Dataset Preprocessing}

\begin{figure}
\centering
\subfloat[Inter-domain translation]{\begin{minipage}{0.40\linewidth}
  \includegraphics[width=0.8\linewidth]{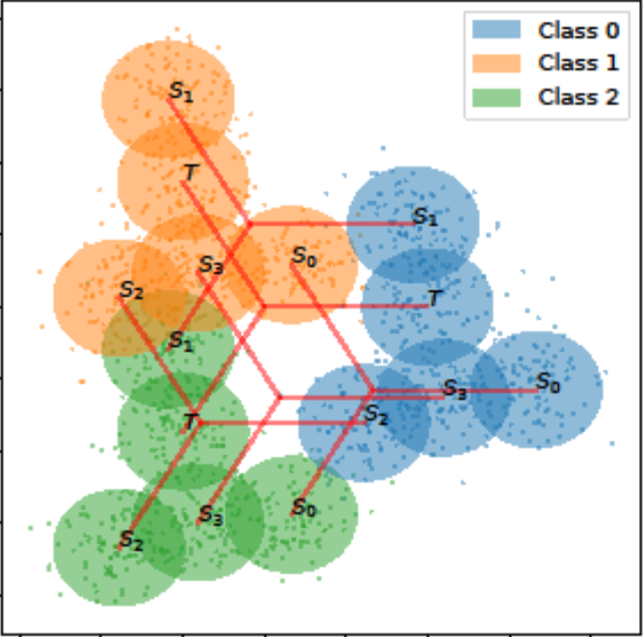}
\end{minipage}}
\hfil
\subfloat[Intra-domain translation]{\begin{minipage}{0.40\linewidth}
  \includegraphics[width=0.8\linewidth]{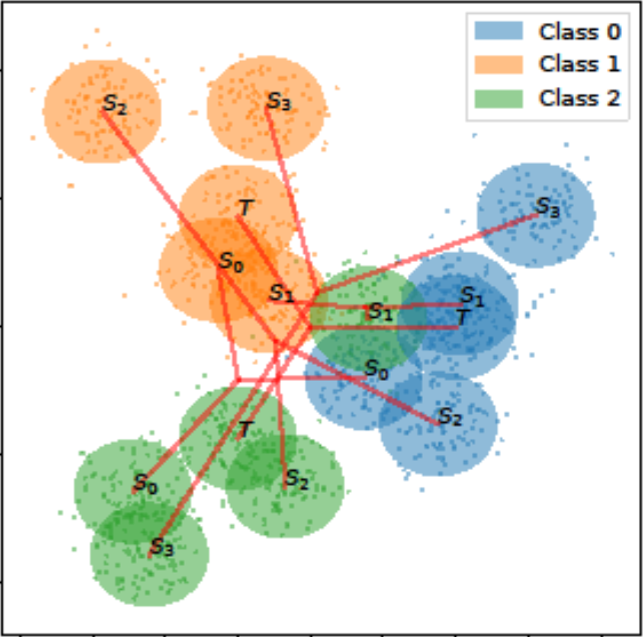}
\end{minipage}}
\hfil
\subfloat[1 GMM]{\begin{minipage}{0.40\linewidth}
  \includegraphics[width=0.8\linewidth]{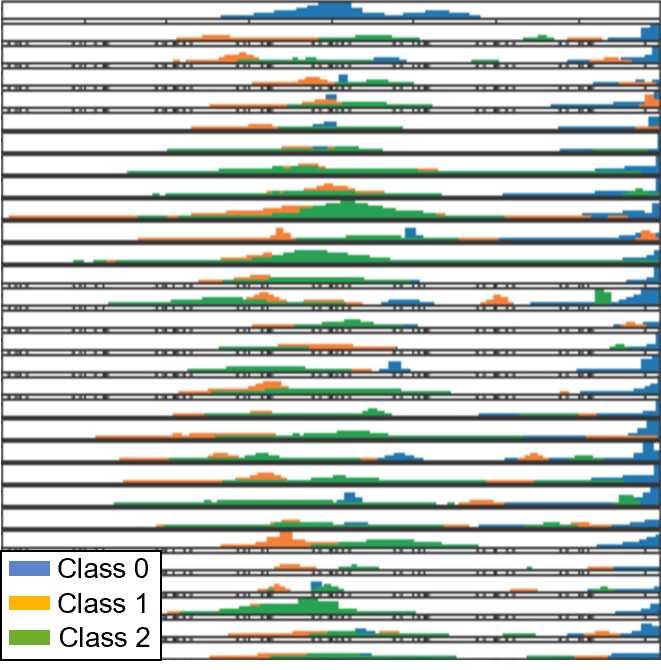}
\end{minipage}}
\hfil
\subfloat[3 GMM]{\begin{minipage}{0.40\linewidth}
\includegraphics[width=0.8\linewidth]{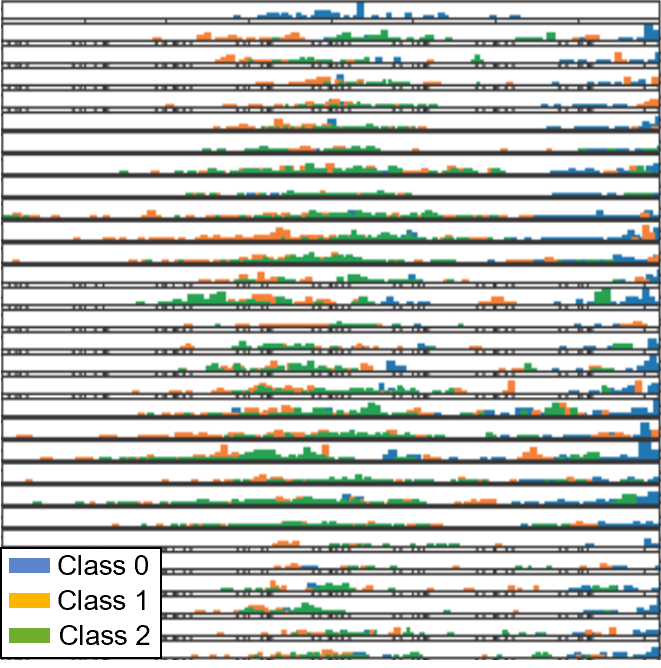}
\end{minipage}}
\caption[Synthetic time series data generated from two sine waves of different frequencies]{SW synthetic data are generated by summing sine waves of different frequencies for each domain and class label. Each source domain is generated with a) inter-domain or b) intra-domain translation or rotation shifts. GMM synthetic data are generated by sampling from from 1, 2, or 3 Gaussians for each domain, dimension, and class label.}
\label{fig:synthetic_data}
\end{figure}

Figure~\ref{fig:synthetic_data} illustrates the synthetic datasets that were used for our experiments.
The synthetic datasets were generated with 12 domains and 3 classes for each domain. The 12 domains allows for $n \in \{2, 4, 6, 8, 10\}$ with exactly 3 random sets of source domains for the largest number of source domains $n=10$. The time series signals were generated at 250 Hz with a window length of 0.2 seconds, yielding a window of 50 samples. Each sine wave had uniform amplitude and no phase shift.

The UCI HAR dataset contains accelerometer $x$, $y$, and $z$, gyroscope $x$, $y$, and $z$, and estimated body acceleration $x$, $y$, and $z$ for 30 participants \cite{anguita2013public}. This data was collected at 50 Hz and is segmented into 2.56 second windows (i.e., each window contains 128 time steps). This dataset contains 6 activity labels: walking, walking upstairs, walking downstairs, sitting, standing, and laying.

The UCI HHAR dataset contains both accelerometer and gyroscope data \cite{stisen2015smartdevices}. However, the authors only utilize one of these sensor modalities at a time and find accelerometer data to be the superior choice for human activity recognition. Thus, on this dataset, we similarly use the three-axis accelerometer data in our experiments. We include the data from the 31 participants carrying smartphones, each of which was sampled at the highest-supported sampling rate, and segment this into windows of 128 time steps. This dataset includes data from the following activities: biking, sitting, standing, walking, walking upstairs, and walking downstairs.

The WISDM AR dataset contains accelerometer $x$, $y$, and $z$ data collected from a large number of participants \cite{kwapisz2011wisdmar}. However, many of these participants have very little data. Thus, we only include data from the 33 participants with sufficient data. The accelerometer data is collected at 20 Hz, and we segment this into non-overlapping windows of 128 time steps. The activity labels for WISDM AR are: walking, jogging, sitting, standing, walking upstairs, and walking downstairs.

The WISDM AT dataset similarly contains accelerometer data \cite{lockhart2011wisdmat}. Like WISDM AR, the amount of data collected from each participant varies widely, so we only include data from the 51 participants with a sufficient amount of labeled data. The accelerometer is sampled at 20 Hz. We segment this data into non-overlapping windows of 128 time steps. WISDM AT contains the following activity labels: walking, jogging, ascending/descending stairs, sitting, standing, and lying down.

The Myo EMG dataset \cite{cote2019emgmyo} contains 8-channel EMG data from a Myo armband collected at 200 Hz while different people performed various hand gestures. This dataset consists of 7 hand gestures: neutral, radial deviation, wrist flexion, ulnar deviation, wrist extension, hand close, and hand open. Data was collected from 40 people, but the authors of the dataset proposed pre-training on the data from the first 22 participants, so they only provide only training sets for these participants. Thus, we include all participants as potential source domains but only include the later 18 participants as target domains. As in their paper, we use 260ms windows (i.e., 52 samples) with an overlap of 235 ms.

The NinaPro Myo dataset consists of data from the NinaPro DB5 Exercise 2 dataset \cite{pizzolato2017ninaprodb5} processed similar to the Myo EMG dataset to better align with the problem setup in their paper \cite{cote2019emgmyo}. The same subset of classes are used as in Myo EMG and only the 8-channel EMG signals from the lower Myo armband. Additionally, while the authors proposed an electrode shifting/rotation mechanism to better align data across participants, we found this additional procedure to be unnecessary for our domain adaptation method. Thus, we perform no such electrode shifting. This dataset consists of data from 10 participants at 200 Hz, and we use the same window size and overlap as the Myo EMG dataset. To not have overlap between data from gestures in the training and testing sets, we select the first 5 repetitions of each gesture as the training data and the final repetition as the test data, which gives approximately an 80\%-20\% train-test split.

For each HAR dataset, the data from each participant was split into 80\% training set and 20\% testing set, with the training set similarly split into training and validation sets. The train-test splits for the EMG datasets were described above, and the training set of each was further split into training and validation sets of 80\% and 20\% respectively. After the selection of source and target domains for each experiment, we select the corresponding training, validation, and test sets for each participant. Data are normalized to have zero mean and unit variance with statistics computed on only the training set. The model selected for evaluation is the checkpoint that performs best on the validation set.

\begin{table*}
\centering
\caption[Ablation study of CALDA instantiations that include the target-domain data via pseudo labeling on synthetic datasets]{Ablation study of CALDA instantiations that include the target-domain data via pseudo labeling on synthetic datasets.}
\label{table:ablation_pseudo_synthetic_shift2}
\begin{scriptsize}
{\renewcommand{\arraystretch}{1.4}
\begin{tabular}{ccccccc}
\toprule
Dataset & \textit{CALDA-In,R,P} & \textit{CALDA-In,H,P} & \textit{CALDA-Any,R,P} & \textit{CALDA-Any,H,P} & \textit{CALDA-XS,R,P} & \textit{CALDA-XS,H,P} \\
\midrule
Synth InterT 10 & 72.1 $\pm$ 12.8 & 72.7 $\pm$ 14.1 & 71.5 $\pm$ 13.6 & 71.1 $\pm$ 15.1 & 70.7 $\pm$ 13.8 & 72.4 $\pm$ 13.7 \\
Synth InterR 1.0 & 69.7 $\pm$ 11.6 & 70.1 $\pm$ 10.1 & 65.4 $\pm$ 10.4 & 77.9 $\pm$ 9.2 & 65.2 $\pm$ 10.7 & 76.6 $\pm$ 9.4 \\
Synth IntraT 10 & 66.7 $\pm$ 6.4 & 68.4 $\pm$ 7.0 & 71.8 $\pm$ 7.5 & 73.7 $\pm$ 11.2 & 69.4 $\pm$ 7.3 & 75.0 $\pm$ 8.9 \\
Synth IntraR 1.0 & 77.7 $\pm$ 6.1 & 77.6 $\pm$ 7.5 & 77.9 $\pm$ 6.9 & 75.9 $\pm$ 3.3 & 78.0 $\pm$ 6.4 & 75.6 $\pm$ 4.3 \\
\hline
Average & 71.6 $\pm$ 9.2 & 72.2 $\pm$ 9.7 & 71.7 $\pm$ 9.6 & 74.7 $\pm$ 9.7 & 70.8 $\pm$ 9.6 & 74.9 $\pm$ 9.1 \\
\bottomrule
\end{tabular}

}
\end{scriptsize}
\end{table*}

\begin{table*}
\centering
\caption[Ablation study comparing CALDA with or without an adversary on synthetic datasets]{Ablation study comparing CALDA with or without an adversary on synthetic datasets. Bold denotes highest accuracy in each row.}
\label{table:ablation_noadv_synthetic_shift2}
\begin{scriptsize}
{\renewcommand{\arraystretch}{1.4}
\begin{tabular}{ccccc}
\toprule
Dataset & \textit{CALDA-Any,R,NoAdv} & \textit{CALDA-XS,H,NoAdv} & \textit{CALDA-Any,R} & \textit{CALDA-XS,H} \\
\midrule
Synth InterT 10 & 70.0 $\pm$ 16.4 & 69.3 $\pm$ 14.0 & 69.4 $\pm$ 17.0 & \textbf{70.5 $\pm$ 14.5} \\
Synth InterR 1.0 & 68.8 $\pm$ 12.0 & 76.2 $\pm$ 9.3 & 63.4 $\pm$ 10.2 & \textbf{76.3 $\pm$ 8.9} \\
Synth IntraT 10 & 74.7 $\pm$ 7.6 & 75.5 $\pm$ 8.6 & 75.6 $\pm$ 8.4 & \textbf{76.2 $\pm$ 8.3} \\
Synth IntraR 1.0 & 73.8 $\pm$ 7.8 & 77.2 $\pm$ 5.5 & \textbf{77.5 $\pm$ 7.0} & 76.9 $\pm$ 6.1 \\
\hline
Average & 71.8 $\pm$ 10.9 & 74.6 $\pm$ 9.4 & 71.5 $\pm$ 10.7 & \textbf{75.0 $\pm$ 9.4} \\
\bottomrule
\end{tabular}

}
\end{scriptsize}
\end{table*}

\begin{table*}
\centering
\caption[Comparison of target domain accuracy for single-source domain adaptation using adversarial learning (CoDATS) and adversarial learning with weak supervision (CoDATS-WS).]{Comparison of target classification accuracy (source $\rightarrow$ target, mean $\pm$ std\%) on randomly-chosen single-source domain adaptation problems for each dataset, using adversarial learning (CoDATS) and adversarial learning with weak supervision (CoDATS-WS). Underline denotes highest accuracy in each row.}
\label{table:ssda}
\tabcolsep=0.11cm
\begin{scriptsize}
{\renewcommand{\arraystretch}{1.2}
\begin{tabular}{cccccc}
\toprule
Problem & No Adaptation & \textit{CoDATS} & \textit{CoDATS-WS} & Train on Target \\
\midrule
HAR 2 $\rightarrow$ 11 & \underline{83.3 $\pm$ 0.7} & 74.5 $\pm$ 4.5 & 74.5 $\pm$ 6.0 & 100.0 $\pm$ 0.0 \\
HAR 7 $\rightarrow$ 13 & 89.9 $\pm$ 3.6 & \underline{96.5 $\pm$ 0.7} & \underline{96.5 $\pm$ 0.7} & 100.0 $\pm$ 0.0 \\
HAR 12 $\rightarrow$ 16 & 41.9 $\pm$ 0.0 & \underline{77.5 $\pm$ 0.6} & 75.2 $\pm$ 3.5 & 100.0 $\pm$ 0.0 \\
HAR 12 $\rightarrow$ 18 & 90.0 $\pm$ 1.7 & \underline{100.0 $\pm$ 0.0} & \underline{100.0 $\pm$ 0.0} & 100.0 $\pm$ 0.0 \\
HAR 9 $\rightarrow$ 18 & 31.1 $\pm$ 1.7 & \underline{85.8 $\pm$ 1.7} & 76.7 $\pm$ 6.8 & 100.0 $\pm$ 0.0 \\
HAR 14 $\rightarrow$ 19 & 62.0 $\pm$ 4.3 & 72.2 $\pm$ 27.2 & \underline{98.6 $\pm$ 1.1} & 100.0 $\pm$ 0.0 \\
HAR 18 $\rightarrow$ 23 & \underline{89.3 $\pm$ 5.0} & 86.2 $\pm$ 0.6 & \underline{89.3 $\pm$ 1.1} & 100.0 $\pm$ 0.0 \\
HAR 6 $\rightarrow$ 23 & 52.9 $\pm$ 2.3 & \underline{94.7 $\pm$ 1.1} & 94.2 $\pm$ 1.3 & 100.0 $\pm$ 0.0 \\
HAR 7 $\rightarrow$ 24 & 94.4 $\pm$ 2.7 & \underline{100.0 $\pm$ 0.0} & 99.1 $\pm$ 0.6 & 100.0 $\pm$ 0.0 \\
HAR 17 $\rightarrow$ 25 & 57.3 $\pm$ 5.5 & 96.7 $\pm$ 1.5 & \underline{97.6 $\pm$ 1.0} & 100.0 $\pm$ 0.0 \\
\hdashline
HAR Average & 69.2 $\pm$ 21.8 & 88.4 $\pm$ 10.1 & \underline{90.2 $\pm$ 10.1} & 100.0 $\pm$ 0.0 \\
\hline
HHAR 1 $\rightarrow$ 3 & 77.8 $\pm$ 4.4 & \underline{93.2 $\pm$ 1.6} & 90.8 $\pm$ 2.0 & 99.2 $\pm$ 0.0 \\
HHAR 3 $\rightarrow$ 5 & 68.8 $\pm$ 5.2 & \underline{95.6 $\pm$ 0.9} & 94.3 $\pm$ 1.2 & 99.0 $\pm$ 0.1 \\
HHAR 4 $\rightarrow$ 5 & 60.4 $\pm$ 3.0 & 94.2 $\pm$ 1.1 & \underline{94.7 $\pm$ 0.5} & 99.0 $\pm$ 0.1 \\
HHAR 0 $\rightarrow$ 6 & 33.6 $\pm$ 2.2 & \underline{76.7 $\pm$ 1.5} & 74.2 $\pm$ 1.1 & 98.8 $\pm$ 0.1 \\
HHAR 1 $\rightarrow$ 6 & 72.1 $\pm$ 3.9 & 90.5 $\pm$ 0.7 & \underline{90.8 $\pm$ 0.2} & 98.8 $\pm$ 0.1 \\
HHAR 4 $\rightarrow$ 6 & 48.0 $\pm$ 2.6 & \underline{93.7 $\pm$ 0.4} & 85.3 $\pm$ 10.6 & 98.8 $\pm$ 0.1 \\
HHAR 5 $\rightarrow$ 6 & 65.1 $\pm$ 6.9 & 90.7 $\pm$ 2.3 & \underline{91.7 $\pm$ 0.4} & 98.8 $\pm$ 0.1 \\
HHAR 2 $\rightarrow$ 7 & 49.4 $\pm$ 2.1 & \underline{58.1 $\pm$ 4.5} & 56.6 $\pm$ 3.4 & 98.5 $\pm$ 0.5 \\
HHAR 3 $\rightarrow$ 8 & 77.8 $\pm$ 2.1 & 93.4 $\pm$ 0.4 & \underline{94.3 $\pm$ 1.0} & 99.3 $\pm$ 0.0 \\
HHAR 5 $\rightarrow$ 8 & 95.3 $\pm$ 0.4 & \underline{97.1 $\pm$ 0.3} & 95.8 $\pm$ 0.2 & 99.3 $\pm$ 0.0 \\
\hdashline
HHAR Average & 64.8 $\pm$ 16.9 & \underline{88.3 $\pm$ 11.4} & 86.8 $\pm$ 11.8 & 99.0 $\pm$ 0.3 \\
\hline
WISDM AR 1 $\rightarrow$ 11 & 71.7 $\pm$ 0.0 & 71.7 $\pm$ 0.0 & \underline{93.3 $\pm$ 0.0} & 98.3 $\pm$ 0.0 \\
WISDM AR 3 $\rightarrow$ 11 & 6.7 $\pm$ 4.9 & \underline{47.8 $\pm$ 0.8} & 46.7 $\pm$ 0.0 & 98.3 $\pm$ 0.0 \\
WISDM AR 4 $\rightarrow$ 15 & 78.2 $\pm$ 4.5 & \underline{81.4 $\pm$ 8.9} & 75.6 $\pm$ 6.3 & 100.0 $\pm$ 0.0 \\
WISDM AR 2 $\rightarrow$ 25 & 81.1 $\pm$ 2.8 & 90.6 $\pm$ 1.6 & \underline{97.8 $\pm$ 0.8} & 100.0 $\pm$ 0.0 \\
WISDM AR 25 $\rightarrow$ 29 & 47.1 $\pm$ 8.2 & 74.6 $\pm$ 7.4 & \underline{84.8 $\pm$ 1.8} & 95.7 $\pm$ 0.0 \\
WISDM AR 7 $\rightarrow$ 30 & 62.5 $\pm$ 0.0 & \underline{73.2 $\pm$ 16.2} & 70.2 $\pm$ 9.9 & 100.0 $\pm$ 0.0 \\
WISDM AR 21 $\rightarrow$ 31 & 57.1 $\pm$ 0.0 & 68.6 $\pm$ 4.0 & \underline{92.4 $\pm$ 1.3} & 97.1 $\pm$ 0.0 \\
WISDM AR 2 $\rightarrow$ 32 & 60.1 $\pm$ 9.1 & 67.3 $\pm$ 0.9 & \underline{68.6 $\pm$ 1.6} & 100.0 $\pm$ 0.0 \\
WISDM AR 1 $\rightarrow$ 7 & 68.5 $\pm$ 2.3 & \underline{70.9 $\pm$ 0.0} & 66.1 $\pm$ 6.9 & 96.4 $\pm$ 0.0 \\
WISDM AR 0 $\rightarrow$ 8 & 34.7 $\pm$ 9.3 & 54.0 $\pm$ 15.6 & \underline{62.0 $\pm$ 15.7} & 99.3 $\pm$ 0.9 \\
\hdashline
WISDM AR Average & 56.8 $\pm$ 21.3 & 70.0 $\pm$ 11.6 & \underline{75.8 $\pm$ 15.4} & 98.5 $\pm$ 1.6 \\
\hline
\bottomrule
\end{tabular}
}
\end{scriptsize}
\end{table*}

\section{Additional Experimental Results}

While the key results were presented in the main paper, here we provide a few additional tables that could not be included in the main paper due to space constraints. These further results corroborate our conclusions from the main paper. Table~\ref{table:ablation_pseudo_synthetic_shift2} provides the ablation results when including pseudo-labeled target domain data on the synthetic datasets. Table~\ref{table:ablation_noadv_synthetic_shift2} provides the results of the ablation study without the adversary on the synthetic datasets.

Table~\ref{table:ssda} summarizes target domain classifier performance for single-source CoDATS domain adaptation. In contrast,
Tables 4 through 8
show the results comparing CALDA with baseline methods when varying the number of source domains $n$ for MS-UDA and the synthetic datasets.
Similarly, Tables~\ref{table:results_ws} and \ref{table:results_ws_synthetic_shift2} show the results for MS-UDA with weak supervision. In these tables, CoDATS represents a configuration where adversarial learning (without or with weak supervision) is employed without the use of contrastive learning.
Note that the average performance of single-source adaptation in Table~\ref{table:ssda} is 83.3, in comparison with an average performance for the same datasets of 89.0 when the largest number of sources is used, as shown in Tables~\ref{table:results_full} and~\ref{table:results_ws}.

\begin{table}
\centering
\caption[Performance of CALDA-Any,R on a modified version of WISDM AR. The dataset name indicates the amount of data that was removed to vary domain lengths. For example, the 60-100\% entry indicates that the length of each domain's time series, including the target, is between 60\% and 100\% of the original size. The amount of removed data is chosen randomly between domains.]{Performance of CALDA-Any,R on a modified version of WISDM AR. The dataset name indicates the amount of data that was removed to vary domain lengths. For example, the 60-100\% entry indicates that the length of each domain's time series, including the target, is between 60\% and 100\% of the original size. The amount of removed data is chosen randomly between domains.} \label{table:varyinglengths}
\begin{tabular}{ccc}
\toprule
Dataset & $n$ & \textit{CALDA-Any,R} \\
\midrule
WISDM AR & 2 & \textbf{68.6 $\pm$ 13.3} \\
WISDM AR & 8 & \textbf{77.9 $\pm$ 8.2} \\
WISDM AR & 14 & \textbf{85.6 $\pm$ 4.7} \\
WISDM AR & 20 & \textbf{85.5 $\pm$ 4.8} \\
WISDM AR & 26 & \textbf{83.2 $\pm$ 4.3} \\
\hdashline[0.75pt/3pt]
WISDM AR & Avg & \textbf{80.2 $\pm$ 7.1} \\
\hline
WISDM AR 90-100\% & 2 & \textbf{70.4 $\pm$ 11.8} \\
WISDM AR 90-100\% & 8 & \textbf{80.6 $\pm$ 7.2} \\
WISDM AR 90-100\% & 14 & \textbf{83.3 $\pm$ 5.2} \\
WISDM AR 90-100\% & 20 & \textbf{88.2 $\pm$ 2.6} \\
WISDM AR 90-100\% & 26 & \textbf{83.6 $\pm$ 4.4} \\
\hdashline[0.75pt/3pt]
WISDM AR 90-100\% & Avg & \textbf{81.2 $\pm$ 6.2} \\
\hline
WISDM AR 80-100\% & 2 & \textbf{66.4 $\pm$ 14.0} \\
WISDM AR 80-100\% & 8 & \textbf{76.9 $\pm$ 10.5} \\
WISDM AR 80-100\% & 14 & \textbf{84.4 $\pm$ 6.0} \\
WISDM AR 80-100\% & 20 & \textbf{84.6 $\pm$ 5.1} \\
WISDM AR 80-100\% & 26 & \textbf{86.2 $\pm$ 2.6} \\
\hdashline[0.75pt/3pt]
WISDM AR 80-100\% & Avg & \textbf{79.7 $\pm$ 7.7} \\
\hline
WISDM AR 60-100\% & 2 & \textbf{65.0 $\pm$ 10.5} \\
WISDM AR 60-100\% & 8 & \textbf{74.3 $\pm$ 9.9} \\
WISDM AR 60-100\% & 14 & \textbf{82.9 $\pm$ 5.6} \\
WISDM AR 60-100\% & 20 & \textbf{85.8 $\pm$ 5.1} \\
WISDM AR 60-100\% & 26 & \textbf{85.9 $\pm$ 4.3} \\
\hdashline[0.75pt/3pt]
WISDM AR 60-100\% & Avg & \textbf{78.8 $\pm$ 7.1} \\
\hline
Average &  & \textbf{80.0 $\pm$ 7.0} \\
\bottomrule
\end{tabular}
\end{table}

Another property of the CALDA framework is that because the models do not reason about the length of the time series itself, the algorithms can handle cases where the domains contain series of different lengths. CALDA's CNN includes a global average pooling layer to handle such cases.
We verify this property using data from the WISDM AR dataset in which we remove elements from the end of time series entries. The amount removed is random, leaving a residual percentage shown in Table~\ref{table:varyinglengths} and resulting in varying time series lengths between domains. As Table~\ref{table:varyinglengths} shows, CALDA is able to handle each of these cases. As the range of lengths increases between domains, the adaptation performance does decrease slightly, due to the increased underlying distribution differences between domains.

\begin{table*}
\centering
\caption[Comparison of target domain accuracy of the most-promising CALDA instantiations with baselines on real-world datasets when varying the number of source domains]{Comparison of target domain accuracy of the most-promising CALDA instantiations with baselines on real-world datasets when varying the number of source domains $n$. Bold denotes CALDA outperforming baselines. Underline denotes highest accuracy in each row.} %
\label{table:results_full}
\begin{scriptsize}
{\renewcommand{\arraystretch}{1.4}
\begin{tabular}{cccccccc}
\toprule
Dataset & $n$ & No Adaptation & CAN & CoDATS & \textit{CALDA-Any,R} & \textit{CALDA-XS,H} & Train on Target \\
\midrule
UCI HAR & 2 & 74.5 $\pm$ 12.2 & 84.1 $\pm$ 7.4 & 91.3 $\pm$ 3.6 & \textbf{\underline{92.1 $\pm$ 4.1}} & 91.2 $\pm$ 4.8 & 99.6 $\pm$ 0.1 \\
UCI HAR & 8 & 90.4 $\pm$ 3.0 & 88.3 $\pm$ 4.1 & \underline{93.7 $\pm$ 2.8} & 92.6 $\pm$ 3.0 & 93.4 $\pm$ 2.4 & 99.6 $\pm$ 0.1 \\
UCI HAR & 14 & 92.4 $\pm$ 2.8 & 90.5 $\pm$ 2.9 & 92.9 $\pm$ 3.9 & \textbf{93.5 $\pm$ 2.3} & \textbf{\underline{94.4 $\pm$ 2.6}} & 99.6 $\pm$ 0.1 \\
UCI HAR & 20 & 93.2 $\pm$ 1.6 & 90.5 $\pm$ 2.7 & 92.7 $\pm$ 3.4 & \textbf{93.6 $\pm$ 1.1} & \textbf{\underline{94.1 $\pm$ 1.5}} & 99.6 $\pm$ 0.1 \\
UCI HAR & 26 & 93.4 $\pm$ 1.5 & 91.7 $\pm$ 1.5 & 93.5 $\pm$ 2.8 & \textbf{\underline{94.0 $\pm$ 1.0}} & \textbf{93.8 $\pm$ 1.4} & 99.6 $\pm$ 0.1 \\
\hdashline[0.75pt/3pt]
UCI HAR & Avg & 88.8 $\pm$ 4.2 & 89.0 $\pm$ 3.7 & 92.8 $\pm$ 3.3 & \textbf{93.1 $\pm$ 2.3} & \textbf{\underline{93.4 $\pm$ 2.5}} & 99.6 $\pm$ 0.1 \\
\hline
UCI HHAR & 2 & 68.4 $\pm$ 7.5 & 73.3 $\pm$ 7.1 & 87.6 $\pm$ 5.1 & \textbf{\underline{88.8 $\pm$ 4.4}} & \textbf{88.7 $\pm$ 4.7} & 98.9 $\pm$ 0.2 \\
UCI HHAR & 3 & 74.8 $\pm$ 6.8 & 77.5 $\pm$ 6.3 & 88.6 $\pm$ 4.3 & \textbf{\underline{90.2 $\pm$ 3.8}} & \textbf{89.7 $\pm$ 3.5} & 98.9 $\pm$ 0.2 \\
UCI HHAR & 4 & 77.9 $\pm$ 7.2 & 79.0 $\pm$ 5.0 & 89.4 $\pm$ 3.4 & \textbf{\underline{90.2 $\pm$ 3.6}} & \textbf{89.9 $\pm$ 4.1} & 98.9 $\pm$ 0.2 \\
UCI HHAR & 5 & 80.9 $\pm$ 5.2 & 79.8 $\pm$ 4.3 & 89.0 $\pm$ 3.2 & \textbf{\underline{90.0 $\pm$ 3.4}} & \textbf{89.6 $\pm$ 3.9} & 98.9 $\pm$ 0.2 \\
UCI HHAR & 6 & 82.4 $\pm$ 4.9 & 78.0 $\pm$ 3.6 & 88.6 $\pm$ 3.4 & \textbf{\underline{89.7 $\pm$ 3.4}} & \textbf{89.3 $\pm$ 3.6} & 98.9 $\pm$ 0.2 \\
\hdashline[0.75pt/3pt]
UCI HHAR & Avg & 76.9 $\pm$ 6.3 & 77.5 $\pm$ 5.3 & 88.6 $\pm$ 3.9 & \textbf{\underline{89.8 $\pm$ 3.7}} & \textbf{89.4 $\pm$ 4.0} & 98.9 $\pm$ 0.2 \\
\hline
WISDM AR & 2 & 55.2 $\pm$ 13.0 & 48.8 $\pm$ 10.1 & 65.6 $\pm$ 13.9 & \textbf{68.6 $\pm$ 13.3} & \textbf{\underline{69.9 $\pm$ 12.3}} & 96.5 $\pm$ 0.1 \\
WISDM AR & 8 & 69.6 $\pm$ 8.2 & 57.8 $\pm$ 7.8 & 76.1 $\pm$ 8.8 & \textbf{77.9 $\pm$ 8.2} & \textbf{\underline{79.9 $\pm$ 10.5}} & 96.5 $\pm$ 0.1 \\
WISDM AR & 14 & 77.8 $\pm$ 7.4 & 65.9 $\pm$ 5.9 & 80.8 $\pm$ 8.7 & \textbf{\underline{85.6 $\pm$ 4.7}} & \textbf{84.2 $\pm$ 5.9} & 96.5 $\pm$ 0.1 \\
WISDM AR & 20 & 78.1 $\pm$ 6.6 & 65.6 $\pm$ 7.3 & 82.3 $\pm$ 5.2 & \textbf{85.5 $\pm$ 4.8} & \textbf{\underline{86.3 $\pm$ 5.9}} & 96.5 $\pm$ 0.1 \\
WISDM AR & 26 & 79.7 $\pm$ 4.5 & 68.2 $\pm$ 6.3 & 85.1 $\pm$ 5.2 & 83.2 $\pm$ 4.3 & \textbf{\underline{86.5 $\pm$ 4.7}} & 96.5 $\pm$ 0.1 \\
\hdashline[0.75pt/3pt]
WISDM AR & Avg & 72.1 $\pm$ 8.0 & 61.3 $\pm$ 7.5 & 78.0 $\pm$ 8.4 & \textbf{80.2 $\pm$ 7.1} & \textbf{\underline{81.4 $\pm$ 7.9}} & 96.5 $\pm$ 0.1 \\
\hline
WISDM AT & 2 & 51.8 $\pm$ 15.8 & 50.3 $\pm$ 13.8 & 56.5 $\pm$ 16.9 & \textbf{\underline{57.9 $\pm$ 13.7}} & 55.8 $\pm$ 19.5 & 98.8 $\pm$ 0.1 \\
WISDM AT & 12 & 69.1 $\pm$ 8.9 & 64.6 $\pm$ 13.8 & 68.6 $\pm$ 7.2 & \textbf{\underline{71.6 $\pm$ 11.4}} & \textbf{69.6 $\pm$ 13.2} & 98.8 $\pm$ 0.1 \\
WISDM AT & 22 & 73.3 $\pm$ 3.6 & 67.6 $\pm$ 7.8 & 71.8 $\pm$ 4.4 & \textbf{\underline{76.5 $\pm$ 4.0}} & \textbf{74.7 $\pm$ 4.5} & 98.8 $\pm$ 0.1 \\
WISDM AT & 32 & 75.8 $\pm$ 3.3 & 69.9 $\pm$ 7.5 & 76.2 $\pm$ 2.2 & \textbf{\underline{78.0 $\pm$ 3.7}} & \textbf{76.8 $\pm$ 3.4} & 98.8 $\pm$ 0.1 \\
WISDM AT & 42 & \underline{79.4 $\pm$ 4.0} & 78.4 $\pm$ 5.1 & 75.6 $\pm$ 2.5 & 76.6 $\pm$ 4.8 & 78.4 $\pm$ 2.4 & 98.8 $\pm$ 0.1 \\
\hdashline[0.75pt/3pt]
WISDM AT & Avg & 69.9 $\pm$ 7.1 & 66.2 $\pm$ 9.6 & 69.7 $\pm$ 6.6 & \textbf{\underline{72.1 $\pm$ 7.5}} & \textbf{71.0 $\pm$ 8.6} & 98.8 $\pm$ 0.1 \\
\hline
Myo EMG & 2 & 71.7 $\pm$ 8.7 & 61.4 $\pm$ 14.4 & 77.4 $\pm$ 10.8 & \textbf{\underline{78.7 $\pm$ 11.6}} & \textbf{78.6 $\pm$ 10.9} & 97.7 $\pm$ 0.1 \\
Myo EMG & 10 & 77.3 $\pm$ 4.9 & 73.8 $\pm$ 6.7 & \underline{84.9 $\pm$ 5.3} & 84.5 $\pm$ 5.0 & 82.5 $\pm$ 5.1 & 97.7 $\pm$ 0.1 \\
Myo EMG & 18 & 79.3 $\pm$ 4.0 & 76.0 $\pm$ 4.8 & 80.8 $\pm$ 4.8 & \textbf{\underline{84.4 $\pm$ 4.8}} & \textbf{\underline{84.4 $\pm$ 4.8}} & 97.7 $\pm$ 0.1 \\
Myo EMG & 26 & 77.8 $\pm$ 4.3 & 79.1 $\pm$ 3.1 & 83.4 $\pm$ 3.6 & \textbf{\underline{85.8 $\pm$ 3.0}} & \textbf{84.6 $\pm$ 3.1} & 97.7 $\pm$ 0.1 \\
Myo EMG & 34 & 80.6 $\pm$ 4.3 & 80.0 $\pm$ 2.6 & 84.5 $\pm$ 2.5 & \textbf{85.3 $\pm$ 3.5} & \textbf{\underline{86.3 $\pm$ 2.5}} & 97.7 $\pm$ 0.1 \\
\hdashline[0.75pt/3pt]
Myo EMG & Avg & 77.4 $\pm$ 5.2 & 74.1 $\pm$ 6.3 & 82.2 $\pm$ 5.4 & \textbf{\underline{83.8 $\pm$ 5.6}} & \textbf{83.3 $\pm$ 5.3} & 97.7 $\pm$ 0.1 \\
\hline
NinaPro Myo & 2 & 48.4 $\pm$ 5.8 & 52.0 $\pm$ 3.4 & 52.7 $\pm$ 5.6 & \textbf{\underline{54.4 $\pm$ 5.1}} & \textbf{53.2 $\pm$ 4.7} & 77.8 $\pm$ 1.3 \\
NinaPro Myo & 4 & 54.8 $\pm$ 4.3 & 56.9 $\pm$ 3.7 & 55.3 $\pm$ 6.3 & \textbf{\underline{58.9 $\pm$ 3.7}} & \textbf{57.0 $\pm$ 4.6} & 77.8 $\pm$ 1.3 \\
NinaPro Myo & 6 & 57.1 $\pm$ 2.6 & \underline{58.4 $\pm$ 3.1} & 57.1 $\pm$ 5.6 & \textbf{\underline{58.4 $\pm$ 3.6}} & 56.9 $\pm$ 3.7 & 77.8 $\pm$ 1.3 \\
NinaPro Myo & 8 & 58.9 $\pm$ 1.6 & \underline{60.2 $\pm$ 2.8} & 58.3 $\pm$ 2.7 & 59.0 $\pm$ 2.9 & 56.9 $\pm$ 2.7 & 77.8 $\pm$ 1.3 \\
\hdashline[0.75pt/3pt]
NinaPro Myo & Avg & 54.8 $\pm$ 3.6 & 56.8 $\pm$ 3.2 & 55.9 $\pm$ 5.0 & \textbf{\underline{57.7 $\pm$ 3.8}} & 56.0 $\pm$ 3.9 & 77.8 $\pm$ 1.3 \\
\hline
Average &  & 73.9 $\pm$ 5.8 & 71.3 $\pm$ 6.0 & 78.6 $\pm$ 5.4 & \textbf{\underline{80.2 $\pm$ 5.0}} & \textbf{79.9 $\pm$ 5.4} & 94.9 $\pm$ 0.3 \\
\bottomrule
\end{tabular}

}
\end{scriptsize}
\end{table*}

\begin{table*}
\centering
\caption[Comparison of target domain accuracy of the most-promising CALDA instantiations with baselines on synthetic SW datasets when varying the number of source domains]{Comparison of target domain accuracy of the most-promising CALDA instantiations with baselines on synthetic SW datasets when varying the number of source domains $n$. Bold denotes CALDA outperforming baselines. Underline denotes highest accuracy in each row.} %
\label{table:results_full_synthetic_shift2}
\begin{scriptsize}
{\renewcommand{\arraystretch}{1.4}
\begin{tabular}{cccccccc}
\toprule
Dataset & $n$ & No Adaptation & CAN & CoDATS & \textit{CALDA-Any,R} & \textit{CALDA-XS,H} & Train on Target \\
\midrule
Synth InterT 10 & 2 & 54.9 $\pm$ 23.5 & \underline{61.2 $\pm$ 12.7} & 57.0 $\pm$ 23.3 & 56.8 $\pm$ 22.0 & 57.1 $\pm$ 20.0 & 93.4 $\pm$ 0.2 \\
Synth InterT 10 & 4 & 56.9 $\pm$ 15.5 & \underline{78.7 $\pm$ 18.9} & 66.3 $\pm$ 18.6 & 59.5 $\pm$ 17.5 & 61.8 $\pm$ 19.2 & 93.4 $\pm$ 0.2 \\
Synth InterT 10 & 6 & 62.3 $\pm$ 18.9 & 77.0 $\pm$ 10.9 & 75.5 $\pm$ 17.6 & 73.1 $\pm$ 24.9 & \textbf{\underline{77.3 $\pm$ 12.0}} & 93.4 $\pm$ 0.2 \\
Synth InterT 10 & 8 & 73.4 $\pm$ 9.6 & 78.3 $\pm$ 8.7 & 75.9 $\pm$ 9.4 & \textbf{\underline{78.6 $\pm$ 9.5}} & 77.8 $\pm$ 9.4 & 93.4 $\pm$ 0.2 \\
Synth InterT 10 & 10 & 65.4 $\pm$ 26.4 & 76.9 $\pm$ 10.2 & 60.2 $\pm$ 12.1 & \textbf{\underline{79.2 $\pm$ 10.9}} & \textbf{78.6 $\pm$ 11.9} & 93.4 $\pm$ 0.2 \\
\hdashline[0.75pt/3pt]
Synth InterT 10 & Avg & 62.6 $\pm$ 18.8 & \underline{74.4 $\pm$ 12.3} & 67.0 $\pm$ 16.2 & 69.4 $\pm$ 17.0 & 70.5 $\pm$ 14.5 & 93.4 $\pm$ 0.2 \\
\hline
Synth InterR 1.0 & 2 & 45.6 $\pm$ 6.7 & \underline{80.2 $\pm$ 2.7} & 51.8 $\pm$ 18.7 & 54.6 $\pm$ 14.7 & 58.0 $\pm$ 16.1 & 94.0 $\pm$ 0.0 \\
Synth InterR 1.0 & 4 & 59.0 $\pm$ 13.5 & 71.4 $\pm$ 22.4 & 55.0 $\pm$ 11.7 & 64.3 $\pm$ 20.1 & \textbf{\underline{72.3 $\pm$ 21.5}} & 94.0 $\pm$ 0.0 \\
Synth InterR 1.0 & 6 & 54.9 $\pm$ 15.5 & 82.4 $\pm$ 6.7 & 63.3 $\pm$ 10.2 & 68.5 $\pm$ 9.8 & \textbf{\underline{83.7 $\pm$ 4.0}} & 94.0 $\pm$ 0.0 \\
Synth InterR 1.0 & 8 & 52.8 $\pm$ 2.5 & 80.5 $\pm$ 1.7 & 56.0 $\pm$ 1.9 & 64.9 $\pm$ 4.6 & \textbf{\underline{83.5 $\pm$ 2.5}} & 94.0 $\pm$ 0.0 \\
Synth InterR 1.0 & 10 & 49.6 $\pm$ 0.9 & 81.2 $\pm$ 0.2 & 54.9 $\pm$ 1.4 & 64.6 $\pm$ 2.0 & \textbf{\underline{84.1 $\pm$ 0.5}} & 94.0 $\pm$ 0.0 \\
\hdashline[0.75pt/3pt]
Synth InterR 1.0 & Avg & 52.4 $\pm$ 7.8 & \underline{79.2 $\pm$ 6.8} & 56.2 $\pm$ 8.8 & 63.4 $\pm$ 10.2 & 76.3 $\pm$ 8.9 & 94.0 $\pm$ 0.0 \\
\hline
Synth IntraT 10 & 2 & 79.8 $\pm$ 4.4 & \underline{91.7 $\pm$ 0.0} & 80.0 $\pm$ 11.5 & 82.0 $\pm$ 5.7 & 80.7 $\pm$ 4.0 & 93.7 $\pm$ 0.2 \\
Synth IntraT 10 & 4 & 69.0 $\pm$ 18.6 & \underline{81.1 $\pm$ 8.6} & 71.8 $\pm$ 9.2 & 71.2 $\pm$ 14.8 & 71.5 $\pm$ 15.6 & 93.7 $\pm$ 0.2 \\
Synth IntraT 10 & 6 & 73.0 $\pm$ 7.5 & \underline{85.2 $\pm$ 7.2} & 61.1 $\pm$ 8.4 & 73.7 $\pm$ 8.1 & 73.9 $\pm$ 10.0 & 93.7 $\pm$ 0.2 \\
Synth IntraT 10 & 8 & 65.0 $\pm$ 5.8 & \underline{83.8 $\pm$ 6.8} & 60.9 $\pm$ 12.6 & 74.5 $\pm$ 4.8 & 79.0 $\pm$ 4.7 & 93.7 $\pm$ 0.2 \\
Synth IntraT 10 & 10 & 66.0 $\pm$ 6.3 & \underline{86.7 $\pm$ 4.1} & 76.7 $\pm$ 2.8 & 76.6 $\pm$ 8.5 & 75.7 $\pm$ 7.0 & 93.7 $\pm$ 0.2 \\
\hdashline[0.75pt/3pt]
Synth IntraT 10 & Avg & 70.6 $\pm$ 8.5 & \underline{85.7 $\pm$ 5.3} & 70.1 $\pm$ 8.9 & 75.6 $\pm$ 8.4 & 76.2 $\pm$ 8.3 & 93.7 $\pm$ 0.2 \\
\hline
Synth IntraR 1.0 & 2 & 70.6 $\pm$ 14.1 & \underline{75.2 $\pm$ 13.1} & 66.0 $\pm$ 15.2 & 69.6 $\pm$ 17.7 & 72.7 $\pm$ 12.4 & 93.6 $\pm$ 0.2 \\
Synth IntraR 1.0 & 4 & 68.1 $\pm$ 3.7 & 73.6 $\pm$ 4.1 & 63.7 $\pm$ 11.1 & \textbf{77.4 $\pm$ 5.5} & \textbf{\underline{81.5 $\pm$ 4.3}} & 93.6 $\pm$ 0.2 \\
Synth IntraR 1.0 & 6 & 59.2 $\pm$ 9.6 & 70.5 $\pm$ 4.2 & 57.8 $\pm$ 9.5 & \textbf{\underline{79.1 $\pm$ 4.4}} & \textbf{76.3 $\pm$ 7.2} & 93.6 $\pm$ 0.2 \\
Synth IntraR 1.0 & 8 & 58.2 $\pm$ 10.9 & 74.5 $\pm$ 5.2 & 61.0 $\pm$ 4.4 & \textbf{\underline{81.7 $\pm$ 4.6}} & \textbf{81.0 $\pm$ 2.7} & 93.6 $\pm$ 0.2 \\
Synth IntraR 1.0 & 10 & 60.4 $\pm$ 6.8 & 73.1 $\pm$ 5.4 & 61.2 $\pm$ 1.8 & \textbf{\underline{79.5 $\pm$ 2.8}} & \textbf{73.1 $\pm$ 3.9} & 93.6 $\pm$ 0.2 \\
\hdashline[0.75pt/3pt]
Synth IntraR 1.0 & Avg & 63.3 $\pm$ 9.0 & 73.4 $\pm$ 6.4 & 61.9 $\pm$ 8.4 & \textbf{\underline{77.5 $\pm$ 7.0}} & \textbf{76.9 $\pm$ 6.1} & 93.6 $\pm$ 0.2 \\
\hline
Average &  & 62.2 $\pm$ 11.0 & \underline{78.2 $\pm$ 7.7} & 63.8 $\pm$ 10.6 & 71.5 $\pm$ 10.7 & 75.0 $\pm$ 9.4 & 93.7 $\pm$ 0.2 \\
\bottomrule
\end{tabular}

}
\end{scriptsize}
\end{table*}

\begin{table*}
\centering
\label{table:results_1gmm}
\caption[Comparison of target domain accuracy of the most-promising CALDA instantiations with baselines on synthetic 1GMM datasets when varying the number of source domains]{Comparison of target domain accuracy of the most-promising CALDA instantiations with baselines on synthetic 1GMM datasets when varying the number of source domains $n$. Bold denotes CALDA outperforming baselines. Underline denotes highest accuracy in each row.}
\begin{scriptsize}
{\renewcommand{\arraystretch}{1.4}
\begin{tabular}{cccccccc}
\toprule
Dataset & $n$ & No Adaptation & CAN & CoDATS & \textit{CALDA-Any,R} & \textit{CALDA-XS,H} & Train on Target \\
\midrule
Synth HAR & 2 & 81.5 $\pm$ 5.8 & 87.7 $\pm$ 5.0 & 91.1 $\pm$ 4.8 & 91.6 $\pm$ 5.0 & \textbf{92.5 $\pm$ 4.9} & 98.4 $\pm$ 0.5 \\
Synth HAR & 8 & 88.2 $\pm$ 3.3 & 89.0 $\pm$ 4.7 & 90.2 $\pm$ 4.2 & \textbf{92.9 $\pm$ 2.8} & 89.8 $\pm$ 2.4 & 98.4 $\pm$ 0.5 \\
Synth HAR & 14 & 89.2 $\pm$ 3.7 & 90.2 $\pm$ 5.5 & \textbf{92.3 $\pm$ 1.8} & 92.1 $\pm$ 4.2 & 91.2 $\pm$ 4.1 & 98.4 $\pm$ 0.5 \\
Synth HAR & 20 & 92.2 $\pm$ 2.0 & 91.4 $\pm$ 2.9 & 92.1 $\pm$ 1.5 & 92.3 $\pm$ 3.1 & \textbf{92.4 $\pm$ 2.6} & 98.4 $\pm$ 0.5 \\
Synth HAR & 26 & 91.4 $\pm$ 2.0 & \textbf{92.4 $\pm$ 3.1} & 91.4 $\pm$ 1.5 & 92.3 $\pm$ 3.6 & 91.4 $\pm$ 2.8 & 98.4 $\pm$ 0.5 \\
\hline
Average &  & 88.5 $\pm$ 3.4 & 90.1 $\pm$ 4.2 & 91.4 $\pm$ 2.8 & \textbf{92.3 $\pm$ 3.7} & 91.4 $\pm$ 3.4 & 98.4 $\pm$ 0.5 \\
\bottomrule
\end{tabular}

}
\end{scriptsize}
\end{table*}

\begin{table*}
\centering
\label{table:results_2gmm}
\caption[Comparison of target domain accuracy of the most-promising CALDA instantiations with baselines on synthetic 2GMM datasets when varying the number of source domains]{Comparison of target domain accuracy of the most-promising CALDA instantiations with baselines on synthetic 2GMM datasets when varying the number of source domains $n$. Bold denotes CALDA outperforming baselines. Underline denotes highest accuracy in each row.}
\begin{scriptsize}
{\renewcommand{\arraystretch}{1.4}
\begin{tabular}{cccccccc}
\toprule
Dataset & $n$ & No Adaptation & CAN & CoDATS & \textit{CALDA-Any,R} & \textit{CALDA-XS,H} & Train on Target \\
\midrule
Synth HAR GMM & 2 & 71.1 $\pm$ 8.9 & 80.6 $\pm$ 8.9 & 80.0 $\pm$ 19.6 & 82.2 $\pm$ 16.3 & \textbf{83.3 $\pm$ 13.6} & 100.0 $\pm$ 0.0 \\
Synth HAR GMM & 8 & 82.2 $\pm$ 10.3 & 91.1 $\pm$ 4.7 & 88.9 $\pm$ 11.0 & 92.2 $\pm$ 6.3 & \textbf{95.6 $\pm$ 4.7} & 100.0 $\pm$ 0.0 \\
Synth HAR GMM & 14 & 90.0 $\pm$ 7.1 & 92.8 $\pm$ 4.7 & 95.6 $\pm$ 4.7 & \textbf{96.7 $\pm$ 3.1} & 95.6 $\pm$ 1.6 & 100.0 $\pm$ 0.0 \\
Synth HAR GMM & 20 & 93.3 $\pm$ 3.9 & 90.0 $\pm$ 7.4 & 95.0 $\pm$ 2.9 & 96.7 $\pm$ 0.0 & \textbf{97.2 $\pm$ 2.1} & 100.0 $\pm$ 0.0 \\
Synth HAR GMM & 26 & 95.0 $\pm$ 2.4 & 92.8 $\pm$ 2.9 & 99.4 $\pm$ 0.8 & \textbf{100.0 $\pm$ 0.0} & 98.3 $\pm$ 1.4 & 100.0 $\pm$ 0.0 \\
\hline
Average &  & 86.3 $\pm$ 6.5 & 89.4 $\pm$ 5.7 & 91.8 $\pm$ 7.8 & 93.6 $\pm$ 5.1 & \textbf{94.0 $\pm$ 4.7} & 100.0 $\pm$ 0.0 \\
\bottomrule
\end{tabular}

}
\end{scriptsize}
\end{table*}

\begin{table*}
\centering
\label{table:results_3gmm}
\caption[Comparison of target domain accuracy of the most-promising CALDA instantiations with baselines on synthetic 3GMM datasets when varying the number of source domains]{Comparison of target domain accuracy of the most-promising CALDA instantiations with baselines on synthetic 3GMM datasets when varying the number of source domains $n$. Bold denotes CALDA outperforming baselines. Underline denotes highest accuracy in each row.}
\begin{scriptsize}
{\renewcommand{\arraystretch}{1.4}
\begin{tabular}{cccccccc}
\toprule
Dataset & $n$ & No Adaptation & CAN & CoDATS & \textit{CALDA-Any,R} & \textit{CALDA-XS,H} & Train on Target \\
\midrule
Synth HAR GMM 3 & 2 & 67.2 $\pm$ 9.5 & 77.2 $\pm$ 11.6 & 78.9 $\pm$ 15.3 & 75.0 $\pm$ 16.7 & \textbf{81.7 $\pm$ 11.8} & 100.0 $\pm$ 0.0 \\
Synth HAR GMM 3 & 8 & 74.4 $\pm$ 10.4 & 84.4 $\pm$ 8.2 & 88.9 $\pm$ 14.1 & \textbf{93.3 $\pm$ 4.7} & 90.6 $\pm$ 10.2 & 100.0 $\pm$ 0.0 \\
Synth HAR GMM 3 & 14 & 83.9 $\pm$ 10.4 & 85.0 $\pm$ 7.4 & 96.7 $\pm$ 4.7 & \textbf{97.2 $\pm$ 3.9} & 96.1 $\pm$ 5.5 & 100.0 $\pm$ 0.0 \\
Synth HAR GMM 3 & 20 & 87.2 $\pm$ 8.6 & 85.0 $\pm$ 8.4 & 95.6 $\pm$ 6.3 & 95.0 $\pm$ 5.5 & \textbf{96.1 $\pm$ 4.5} & 100.0 $\pm$ 0.0 \\
Synth HAR GMM 3 & 26 & 87.8 $\pm$ 6.1 & 87.2 $\pm$ 9.8 & \textbf{96.7 $\pm$ 0.0} & \textbf{96.7 $\pm$ 3.7} & \textbf{96.7 $\pm$ 3.9} & 100.0 $\pm$ 0.0 \\
\hline
Average &  & 80.1 $\pm$ 9.0 & 83.8 $\pm$ 9.1 & 91.3 $\pm$ 8.1 & 91.4 $\pm$ 6.9 & \textbf{92.2 $\pm$ 7.2} & 100.0 $\pm$ 0.0 \\
\bottomrule
\end{tabular}

}
\end{scriptsize}
\end{table*}

\begin{table*}
\centering
\caption[Comparison of target domain accuracy for domain adaptation methods utilizing weak supervision on real-world datasets]{Comparison of target domain accuracy for domain adaptation methods utilizing weak supervision on real-world datasets. Bold denotes CALDA outperforming baselines. Underline denotes highest accuracy in each row.}
\label{table:results_ws}
\begin{scriptsize}
{\renewcommand{\arraystretch}{1.4}
\begin{tabular}{ccccccc}
\toprule
Dataset & $n$ & No Adaptation & CoDATS-WS & \textit{CALDA-Any,R,WS} & \textit{CALDA-XS,H,WS} & Train on Target \\
\midrule
UCI HAR & 2 & 74.5 $\pm$ 12.2 & 91.5 $\pm$ 3.7 & \textbf{\underline{91.8 $\pm$ 4.4}} & \textbf{91.6 $\pm$ 4.4} & 99.6 $\pm$ 0.1 \\
UCI HAR & 8 & 90.4 $\pm$ 3.0 & 92.1 $\pm$ 3.4 & \textbf{94.5 $\pm$ 2.2} & \textbf{\underline{95.4 $\pm$ 2.0}} & 99.6 $\pm$ 0.1 \\
UCI HAR & 14 & 92.4 $\pm$ 2.8 & 94.4 $\pm$ 2.9 & \textbf{95.6 $\pm$ 1.0} & \textbf{\underline{97.0 $\pm$ 1.4}} & 99.6 $\pm$ 0.1 \\
UCI HAR & 20 & 93.2 $\pm$ 1.6 & 93.2 $\pm$ 3.0 & \textbf{95.7 $\pm$ 1.0} & \textbf{\underline{96.5 $\pm$ 1.8}} & 99.6 $\pm$ 0.1 \\
UCI HAR & 26 & 93.4 $\pm$ 1.5 & 93.5 $\pm$ 3.2 & \textbf{96.6 $\pm$ 0.7} & \textbf{\underline{96.8 $\pm$ 0.8}} & 99.6 $\pm$ 0.1 \\
\hdashline[0.75pt/3pt]
UCI HAR & Avg & 88.8 $\pm$ 4.2 & 92.9 $\pm$ 3.2 & \textbf{94.8 $\pm$ 1.8} & \textbf{\underline{95.5 $\pm$ 2.1}} & 99.6 $\pm$ 0.1 \\
\hline
UCI HHAR & 2 & 68.4 $\pm$ 7.5 & 86.7 $\pm$ 5.8 & \textbf{\underline{89.1 $\pm$ 4.6}} & \textbf{88.1 $\pm$ 5.5} & 98.9 $\pm$ 0.2 \\
UCI HHAR & 3 & 74.8 $\pm$ 6.8 & 88.3 $\pm$ 5.2 & \textbf{\underline{90.8 $\pm$ 3.7}} & \textbf{90.0 $\pm$ 4.2} & 98.9 $\pm$ 0.2 \\
UCI HHAR & 4 & 77.9 $\pm$ 7.2 & 89.0 $\pm$ 4.4 & \textbf{\underline{91.0 $\pm$ 3.5}} & \textbf{90.5 $\pm$ 3.7} & 98.9 $\pm$ 0.2 \\
UCI HHAR & 5 & 80.9 $\pm$ 5.2 & 88.6 $\pm$ 3.9 & \textbf{\underline{90.5 $\pm$ 3.4}} & \textbf{90.2 $\pm$ 3.7} & 98.9 $\pm$ 0.2 \\
UCI HHAR & 6 & 82.4 $\pm$ 4.9 & 88.5 $\pm$ 3.5 & \textbf{89.7 $\pm$ 3.8} & \textbf{\underline{90.2 $\pm$ 3.5}} & 98.9 $\pm$ 0.2 \\
\hdashline[0.75pt/3pt]
UCI HHAR & Avg & 76.9 $\pm$ 6.3 & 88.2 $\pm$ 4.6 & \textbf{\underline{90.2 $\pm$ 3.8}} & \textbf{89.8 $\pm$ 4.1} & 98.9 $\pm$ 0.2 \\
\hline
WISDM AR & 2 & 55.2 $\pm$ 13.0 & 77.1 $\pm$ 11.7 & \textbf{\underline{80.6 $\pm$ 8.8}} & \textbf{77.5 $\pm$ 11.8} & 96.5 $\pm$ 0.1 \\
WISDM AR & 8 & 69.6 $\pm$ 8.2 & 85.6 $\pm$ 5.3 & \textbf{\underline{86.0 $\pm$ 7.3}} & 85.0 $\pm$ 8.8 & 96.5 $\pm$ 0.1 \\
WISDM AR & 14 & 77.8 $\pm$ 7.4 & \underline{88.6 $\pm$ 5.0} & 86.0 $\pm$ 6.8 & 86.2 $\pm$ 5.0 & 96.5 $\pm$ 0.1 \\
WISDM AR & 20 & 78.1 $\pm$ 6.6 & \underline{88.3 $\pm$ 4.5} & 86.5 $\pm$ 6.6 & 86.7 $\pm$ 5.5 & 96.5 $\pm$ 0.1 \\
WISDM AR & 26 & 79.7 $\pm$ 4.5 & 84.7 $\pm$ 9.5 & \textbf{86.8 $\pm$ 5.0} & \textbf{\underline{87.9 $\pm$ 4.1}} & 96.5 $\pm$ 0.1 \\
\hdashline[0.75pt/3pt]
WISDM AR & Avg & 72.1 $\pm$ 8.0 & 84.9 $\pm$ 7.2 & \textbf{\underline{85.2 $\pm$ 6.9}} & 84.7 $\pm$ 7.0 & 96.5 $\pm$ 0.1 \\
\hline
WISDM AT & 2 & 51.8 $\pm$ 15.8 & 66.8 $\pm$ 13.5 & \textbf{67.8 $\pm$ 15.6} & \textbf{\underline{68.5 $\pm$ 17.0}} & 98.8 $\pm$ 0.1 \\
WISDM AT & 12 & 69.1 $\pm$ 8.9 & 70.9 $\pm$ 9.1 & \textbf{86.4 $\pm$ 8.0} & \textbf{\underline{87.0 $\pm$ 6.2}} & 98.8 $\pm$ 0.1 \\
WISDM AT & 22 & 73.3 $\pm$ 3.6 & 74.0 $\pm$ 10.1 & \textbf{85.0 $\pm$ 5.9} & \textbf{\underline{86.2 $\pm$ 4.7}} & 98.8 $\pm$ 0.1 \\
WISDM AT & 32 & 75.8 $\pm$ 3.3 & 71.1 $\pm$ 8.4 & \textbf{87.0 $\pm$ 5.0} & \textbf{\underline{88.0 $\pm$ 3.2}} & 98.8 $\pm$ 0.1 \\
WISDM AT & 42 & 79.4 $\pm$ 4.0 & 77.8 $\pm$ 10.2 & \textbf{\underline{88.3 $\pm$ 2.2}} & \textbf{87.0 $\pm$ 4.3} & 98.8 $\pm$ 0.1 \\
\hdashline[0.75pt/3pt]
WISDM AT & Avg & 69.9 $\pm$ 7.1 & 72.1 $\pm$ 10.3 & \textbf{82.9 $\pm$ 7.3} & \textbf{\underline{83.3 $\pm$ 7.1}} & 98.8 $\pm$ 0.1 \\
\hline
Myo EMG & 2 & 71.7 $\pm$ 8.7 & 75.2 $\pm$ 11.6 & \textbf{\underline{79.1 $\pm$ 8.6}} & \textbf{78.8 $\pm$ 9.6} & 97.7 $\pm$ 0.1 \\
Myo EMG & 10 & 77.3 $\pm$ 4.9 & 79.4 $\pm$ 5.2 & \textbf{\underline{87.9 $\pm$ 3.9}} & \textbf{86.7 $\pm$ 5.2} & 97.7 $\pm$ 0.1 \\
Myo EMG & 18 & 79.3 $\pm$ 4.0 & 79.1 $\pm$ 5.5 & \textbf{\underline{87.0 $\pm$ 3.8}} & \textbf{86.1 $\pm$ 4.4} & 97.7 $\pm$ 0.1 \\
Myo EMG & 26 & 77.8 $\pm$ 4.3 & 82.8 $\pm$ 3.4 & \textbf{84.4 $\pm$ 3.6} & \textbf{\underline{84.9 $\pm$ 2.2}} & 97.7 $\pm$ 0.1 \\
Myo EMG & 34 & 80.6 $\pm$ 4.3 & 81.2 $\pm$ 3.0 & \textbf{\underline{87.4 $\pm$ 2.8}} & \textbf{86.8 $\pm$ 1.8} & 97.7 $\pm$ 0.1 \\
\hdashline[0.75pt/3pt]
Myo EMG & Avg & 77.4 $\pm$ 5.2 & 79.5 $\pm$ 5.7 & \textbf{\underline{85.1 $\pm$ 4.6}} & \textbf{84.7 $\pm$ 4.6} & 97.7 $\pm$ 0.1 \\
\hline
NinaPro Myo & 2 & 48.4 $\pm$ 5.8 & 50.6 $\pm$ 5.9 & \textbf{\underline{55.7 $\pm$ 4.7}} & \textbf{51.5 $\pm$ 6.0} & 77.8 $\pm$ 1.3 \\
NinaPro Myo & 4 & 54.8 $\pm$ 4.3 & 56.3 $\pm$ 4.2 & \textbf{\underline{61.1 $\pm$ 4.2}} & \textbf{57.3 $\pm$ 4.2} & 77.8 $\pm$ 1.3 \\
NinaPro Myo & 6 & 57.1 $\pm$ 2.6 & 55.4 $\pm$ 3.8 & \textbf{\underline{58.8 $\pm$ 2.7}} & 56.7 $\pm$ 4.6 & 77.8 $\pm$ 1.3 \\
NinaPro Myo & 8 & 58.9 $\pm$ 1.6 & 57.5 $\pm$ 3.1 & \textbf{\underline{59.5 $\pm$ 3.7}} & 58.4 $\pm$ 3.0 & 77.8 $\pm$ 1.3 \\
\hdashline[0.75pt/3pt]
NinaPro Myo & Avg & 54.8 $\pm$ 3.6 & 54.9 $\pm$ 4.2 & \textbf{\underline{58.8 $\pm$ 3.8}} & \textbf{56.0 $\pm$ 4.5} & 77.8 $\pm$ 1.3 \\
\hline
Average &  & 73.9 $\pm$ 5.8 & 79.6 $\pm$ 5.9 & \textbf{\underline{83.7 $\pm$ 4.7}} & \textbf{83.2 $\pm$ 4.9} & 94.9 $\pm$ 0.3 \\
\bottomrule
\end{tabular}

}
\end{scriptsize}
\end{table*}

\begin{table*}
\centering
\caption[Comparison of target domain accuracy for domain adaptation methods utilizing weak supervision on synthetic SW datasets]{Comparison of target domain accuracy for domain adaptation methods utilizing weak supervision on synthetic SW datasets. Bold denotes CALDA outperforming baselines. Underline denotes highest accuracy in each row.}
\label{table:results_ws_synthetic_shift2}
\begin{scriptsize}
{\renewcommand{\arraystretch}{1.4}
\begin{tabular}{ccccccc}
\toprule
Dataset & $n$ & No Adaptation & CoDATS-WS & \textit{CALDA-Any,R,WS} & \textit{CALDA-XS,H,WS} & Train on Target \\
\midrule
Synth InterT 10 & 2 & 54.9 $\pm$ 23.5 & 60.9 $\pm$ 21.2 & 56.6 $\pm$ 32.1 & \textbf{\underline{74.9 $\pm$ 14.1}} & 93.4 $\pm$ 0.2 \\
Synth InterT 10 & 4 & 56.9 $\pm$ 15.5 & 69.9 $\pm$ 23.1 & \textbf{\underline{82.2 $\pm$ 14.8}} & \textbf{79.4 $\pm$ 16.1} & 93.4 $\pm$ 0.2 \\
Synth InterT 10 & 6 & 62.3 $\pm$ 18.9 & 75.6 $\pm$ 17.4 & \textbf{\underline{83.8 $\pm$ 8.7}} & \textbf{\underline{83.8 $\pm$ 9.5}} & 93.4 $\pm$ 0.2 \\
Synth InterT 10 & 8 & 73.4 $\pm$ 9.6 & 75.0 $\pm$ 15.5 & \textbf{81.4 $\pm$ 13.8} & \textbf{\underline{81.7 $\pm$ 10.4}} & 93.4 $\pm$ 0.2 \\
Synth InterT 10 & 10 & 65.4 $\pm$ 26.4 & 56.3 $\pm$ 20.5 & \textbf{72.9 $\pm$ 20.9} & \textbf{\underline{73.0 $\pm$ 19.0}} & 93.4 $\pm$ 0.2 \\
\hdashline[0.75pt/3pt]
Synth InterT 10 & Avg & 62.6 $\pm$ 18.8 & 67.5 $\pm$ 19.5 & \textbf{75.4 $\pm$ 18.0} & \textbf{\underline{78.6 $\pm$ 13.8}} & 93.4 $\pm$ 0.2 \\
\hline
Synth InterR 1.0 & 2 & 45.6 $\pm$ 6.7 & 47.8 $\pm$ 20.5 & \textbf{54.9 $\pm$ 15.4} & \textbf{\underline{62.1 $\pm$ 16.2}} & 94.0 $\pm$ 0.0 \\
Synth InterR 1.0 & 4 & 59.0 $\pm$ 13.5 & 63.4 $\pm$ 4.9 & \textbf{66.5 $\pm$ 17.5} & \textbf{\underline{73.0 $\pm$ 23.2}} & 94.0 $\pm$ 0.0 \\
Synth InterR 1.0 & 6 & 54.9 $\pm$ 15.5 & 68.3 $\pm$ 10.7 & \textbf{72.2 $\pm$ 14.8} & \textbf{\underline{85.6 $\pm$ 4.2}} & 94.0 $\pm$ 0.0 \\
Synth InterR 1.0 & 8 & 52.8 $\pm$ 2.5 & 56.6 $\pm$ 0.8 & \textbf{63.8 $\pm$ 1.8} & \textbf{\underline{86.3 $\pm$ 2.2}} & 94.0 $\pm$ 0.0 \\
Synth InterR 1.0 & 10 & 49.6 $\pm$ 0.9 & 49.9 $\pm$ 6.8 & \textbf{68.3 $\pm$ 4.8} & \textbf{\underline{84.0 $\pm$ 1.9}} & 94.0 $\pm$ 0.0 \\
\hdashline[0.75pt/3pt]
Synth InterR 1.0 & Avg & 52.4 $\pm$ 7.8 & 57.2 $\pm$ 8.7 & \textbf{65.1 $\pm$ 10.9} & \textbf{\underline{78.2 $\pm$ 9.5}} & 94.0 $\pm$ 0.0 \\
\hline
Synth IntraT 10 & 2 & 79.8 $\pm$ 4.4 & 74.4 $\pm$ 10.0 & \textbf{\underline{86.7 $\pm$ 1.8}} & \textbf{83.1 $\pm$ 0.4} & 93.7 $\pm$ 0.2 \\
Synth IntraT 10 & 4 & 69.0 $\pm$ 18.6 & 58.8 $\pm$ 5.7 & \textbf{\underline{74.1 $\pm$ 18.5}} & \textbf{72.3 $\pm$ 20.0} & 93.7 $\pm$ 0.2 \\
Synth IntraT 10 & 6 & 73.0 $\pm$ 7.5 & 64.3 $\pm$ 10.9 & \textbf{77.3 $\pm$ 7.9} & \textbf{\underline{77.9 $\pm$ 8.4}} & 93.7 $\pm$ 0.2 \\
Synth IntraT 10 & 8 & 65.0 $\pm$ 5.8 & 58.4 $\pm$ 8.2 & \textbf{79.5 $\pm$ 4.8} & \textbf{\underline{81.1 $\pm$ 3.9}} & 93.7 $\pm$ 0.2 \\
Synth IntraT 10 & 10 & 66.0 $\pm$ 6.3 & 61.7 $\pm$ 5.8 & \textbf{74.4 $\pm$ 7.0} & \textbf{\underline{76.9 $\pm$ 2.7}} & 93.7 $\pm$ 0.2 \\
\hdashline[0.75pt/3pt]
Synth IntraT 10 & Avg & 70.6 $\pm$ 8.5 & 63.5 $\pm$ 8.1 & \textbf{\underline{78.4 $\pm$ 8.0}} & \textbf{78.3 $\pm$ 7.1} & 93.7 $\pm$ 0.2 \\
\hline
Synth IntraR 1.0 & 2 & 70.6 $\pm$ 14.1 & 60.8 $\pm$ 22.3 & \textbf{71.6 $\pm$ 19.8} & \textbf{\underline{73.0 $\pm$ 14.7}} & 93.6 $\pm$ 0.2 \\
Synth IntraR 1.0 & 4 & 68.1 $\pm$ 3.7 & 56.6 $\pm$ 7.1 & \textbf{82.3 $\pm$ 6.8} & \textbf{\underline{82.5 $\pm$ 4.6}} & 93.6 $\pm$ 0.2 \\
Synth IntraR 1.0 & 6 & 59.2 $\pm$ 9.6 & 52.0 $\pm$ 7.2 & \textbf{78.8 $\pm$ 8.4} & \textbf{\underline{79.8 $\pm$ 8.0}} & 93.6 $\pm$ 0.2 \\
Synth IntraR 1.0 & 8 & 58.2 $\pm$ 10.9 & 64.5 $\pm$ 3.5 & \textbf{\underline{84.7 $\pm$ 2.8}} & \textbf{83.5 $\pm$ 0.7} & 93.6 $\pm$ 0.2 \\
Synth IntraR 1.0 & 10 & 60.4 $\pm$ 6.8 & 69.7 $\pm$ 10.8 & \textbf{78.0 $\pm$ 2.6} & \textbf{\underline{80.6 $\pm$ 3.2}} & 93.6 $\pm$ 0.2 \\
\hdashline[0.75pt/3pt]
Synth IntraR 1.0 & Avg & 63.3 $\pm$ 9.0 & 60.7 $\pm$ 10.2 & \textbf{79.1 $\pm$ 8.1} & \textbf{\underline{79.9 $\pm$ 6.3}} & 93.6 $\pm$ 0.2 \\
\hline
Average &  & 62.2 $\pm$ 11.0 & 62.2 $\pm$ 11.6 & \textbf{74.5 $\pm$ 11.3} & \textbf{\underline{78.7 $\pm$ 9.2}} & 93.7 $\pm$ 0.2 \\
\bottomrule
\end{tabular}

}
\end{scriptsize}
\end{table*}

\begin{table*}
\centering
\caption[Comparison of target domain accuracy for domain adaptation methods utilizing weak supervision on GMM synthetic datasets]{Comparison of target domain accuracy for domain adaptation methods utilizing weak supervision on GMM synthetic datasets. Bold denotes CALDA outperforming baselines. Underline denotes highest accuracy in each row.}
\begin{scriptsize}
{\renewcommand{\arraystretch}{1.4}
\begin{tabular}{ccccccc}
\toprule
Dataset & $n$ & No Adaptation & CoDATS-WS & \textit{CALDA-Any,R,WS} & \textit{CALDA-XS,H,WS} & Train on Target \\
\midrule
Synth HAR GMM 1 & 2 & 81.5 $\pm$ 5.8 & 90.9 $\pm$ 4.9 & \textbf{\underline{91.9 $\pm$ 4.6}} & 90.0 $\pm$ 6.2 & 98.4 $\pm$ 0.5 \\
Synth HAR GMM 1 & 8 & 88.2 $\pm$ 3.3 & 89.6 $\pm$ 3.4 & \textbf{\underline{90.7 $\pm$ 2.1}} & \textbf{\underline{90.7 $\pm$ 2.3}} & 98.4 $\pm$ 0.5 \\
Synth HAR GMM 1 & 14 & 89.2 $\pm$ 3.7 & 90.2 $\pm$ 3.1 & \textbf{\underline{93.5 $\pm$ 3.3}} & \textbf{91.5 $\pm$ 3.6} & 98.4 $\pm$ 0.5 \\
Synth HAR GMM 1 & 20 & \underline{92.2 $\pm$ 2.0} & 89.8 $\pm$ 1.2 & 91.5 $\pm$ 1.3 & 90.5 $\pm$ 4.9 & 98.4 $\pm$ 0.5 \\
Synth HAR GMM 1 & 26 & \underline{91.4 $\pm$ 2.0} & \underline{91.4 $\pm$ 2.3} & 91.1 $\pm$ 1.2 & 91.1 $\pm$ 1.5 & 98.4 $\pm$ 0.5 \\
\hdashline[0.75pt/3pt]
Synth HAR GMM 1 & Avg & 88.5 $\pm$ 3.4 & 90.4 $\pm$ 3.0 & \textbf{\underline{91.7 $\pm$ 2.5}} & \textbf{90.8 $\pm$ 3.7} & 98.4 $\pm$ 0.5 \\
\hline
Synth HAR GMM 2 & 2 & 71.1 $\pm$ 8.9 & 77.8 $\pm$ 17.7 & \textbf{79.4 $\pm$ 17.5} & \textbf{\underline{82.8 $\pm$ 17.3}} & 100.0 $\pm$ 0.0 \\
Synth HAR GMM 2 & 8 & 82.2 $\pm$ 10.3 & 87.8 $\pm$ 12.3 & \textbf{\underline{92.8 $\pm$ 10.2}} & \textbf{91.1 $\pm$ 12.6} & 100.0 $\pm$ 0.0 \\
Synth HAR GMM 2 & 14 & 90.0 $\pm$ 7.1 & 87.8 $\pm$ 11.0 & \textbf{93.9 $\pm$ 3.9} & \textbf{\underline{97.8 $\pm$ 1.6}} & 100.0 $\pm$ 0.0 \\
Synth HAR GMM 2 & 20 & 93.3 $\pm$ 3.9 & 89.4 $\pm$ 13.4 & \textbf{96.7 $\pm$ 3.1} & \textbf{\underline{100.0 $\pm$ 0.0}} & 100.0 $\pm$ 0.0 \\
Synth HAR GMM 2 & 26 & 95.0 $\pm$ 2.4 & 89.4 $\pm$ 9.9 & \textbf{\underline{98.9 $\pm$ 1.6}} & \textbf{96.7 $\pm$ 4.7} & 100.0 $\pm$ 0.0 \\
\hdashline[0.75pt/3pt]
Synth HAR GMM 2 & Avg & 86.3 $\pm$ 6.5 & 86.4 $\pm$ 12.9 & \textbf{92.3 $\pm$ 7.3} & \textbf{\underline{93.7 $\pm$ 7.2}} & 100.0 $\pm$ 0.0 \\
\hline
Synth HAR GMM 3 & 2 & 67.2 $\pm$ 9.5 & 78.9 $\pm$ 15.4 & 72.2 $\pm$ 14.4 & \textbf{\underline{80.0 $\pm$ 15.7}} & 100.0 $\pm$ 0.0 \\
Synth HAR GMM 3 & 8 & 74.4 $\pm$ 10.4 & 87.2 $\pm$ 9.8 & 86.1 $\pm$ 11.5 & \textbf{\underline{91.7 $\pm$ 7.9}} & 100.0 $\pm$ 0.0 \\
Synth HAR GMM 3 & 14 & 83.9 $\pm$ 10.4 & \underline{95.6 $\pm$ 4.3} & 90.0 $\pm$ 14.1 & 93.3 $\pm$ 7.9 & 100.0 $\pm$ 0.0 \\
Synth HAR GMM 3 & 20 & 87.2 $\pm$ 8.6 & \underline{91.1 $\pm$ 8.7} & 87.8 $\pm$ 13.4 & 88.9 $\pm$ 12.6 & 100.0 $\pm$ 0.0 \\
Synth HAR GMM 3 & 26 & 87.8 $\pm$ 6.1 & 85.6 $\pm$ 7.6 & 84.4 $\pm$ 14.1 & \textbf{\underline{91.7 $\pm$ 10.2}} & 100.0 $\pm$ 0.0 \\
\hdashline[0.75pt/3pt]
Synth HAR GMM 3 & Avg & 80.1 $\pm$ 9.0 & 87.7 $\pm$ 9.2 & 84.1 $\pm$ 13.5 & \textbf{\underline{89.1 $\pm$ 10.9}} & 100.0 $\pm$ 0.0 \\
\hline
Average &  & 85.0 $\pm$ 6.3 & 88.2 $\pm$ 8.3 & \textbf{89.4 $\pm$ 7.8} & \textbf{\underline{91.2 $\pm$ 7.3}} & 99.5 $\pm$ 0.2 \\
\bottomrule
\end{tabular}

}
\end{scriptsize}
\end{table*}

\section{Training Algorithm}

For added clarity, we present the overall training loop using pseudo code in Algorithm~\ref{alg:trainloop} and the contrastive loss computation in Algorithm~\ref{alg:contrastive}. Note, in practice, we use a vectorized version of Algorithm~\ref{alg:contrastive} on GPUs for computational efficiency.

\begin{algorithm}
\DontPrintSemicolon
    \KwIn{$dataset_{train}, dataset_{valid}, F, C, D, Z, GRL, \lambda_c$}
    \KwOut{checkpoint of best model $F, C, D, Z$ on $dataset_{valid}$}
    \tcp{Mini-batch from shuffled/repeated train data}
    \For{$(s_1, s_2, \dots, t) \in dataset_{train}$}{
        \tcp{Concatenate all $x$ and $d$ values}
        x = concat([$s_1$.x, $s_2$.x, \dots, $t$.x])\;
        d = concat([$s_1$.d, $s_2$.d, \dots, $t$.d])\;
        \tcp{Concatenate $y$, exclude unlabeled target}
        y = concat([$s_1$.y, $s_2$.y, \dots, $s_n$.y])\;

        \tcp{Run input through model}
        $f = F(x)$\;
        $y_{pred} = C(f)$\;
        $d_{pred} = D(GRL(f))$ \tcp{Gradient reversal}
        $z = Z(f)$\;

        \tcp{Task loss, ignoring target}
        $loss_{task}$ = cross\_entropy($y$, $y_{pred}$ excluding target)\;

        \tcp{Domain loss, over both sources and target}
        $loss_{domain}$ = cross\_entropy($d$, $d_{pred}$)\;

        \tcp{Contrastive loss, see Algorithm~\ref{alg:contrastive}}
        $loss_{contrastive}$ = contrastive\_loss($y$, $d$, $y_{pred}$, $z$)\;

        \tcp{Total loss, recall GRL includes $\lambda_d$ weight}
        $loss$ = $loss_{task}$ + $loss_{domain}$ + $\lambda_c \cdot loss_{contrastive}$\;

        \tcp{Compute/apply gradient to update model}
        update($loss$, [$F, C, D, Z$])\;

        \tcp{Store best model on validation set}
        checkpoint($dataset_{valid}$, [$F, C, D, Z$])\;
    }
    \caption{Training Loop}
    \label{alg:trainloop}
\end{algorithm}

\begin{algorithm}
\begin{footnotesize}
\DontPrintSemicolon
\KwIn{$y, d, y_{pred}, z$}
\KwOut{$loss$}
    \tcp{Construct query set}
    \eIf{pseudo labeling}{
        append argmax($y_{pred}$) to $y$\;
        $queries$ = $(z, d, y)$ for each example\;
    }{
        $queries$ = $(z, d, y)$ for each source example\;
    }

    $loss$ = 0\;

    \For{$(z_q, d_q, y_q) \in queries$}{
        \tcp{Construct positive and negative sets}
        \If{within source}{
            $positives$ = $z$ with $d_q$ and $y_q$, exclude $z_q$\;
            $negatives$ = $z$ with $d_q$ and not $y_q$\;
        }
        \ElseIf{any source}{
            $positives$ = $z$ with $y_q$, exclude $z_q$\;
            $negatives$ = $z$ with not $y_q$\;
        }\ElseIf{cross source}{
            $positives$ = $z$ with not $d_q$ and $y_q$\;
            $negatives$ = $z$ with not $d_q$ and not $y_q$\;
        }

        \tcp{Hard or random sampling}
        \If{hard sampling}{
            sort $positives$ descending by cross\_entropy($y$ for each positive, $y_{pred}$ for each positive)\;
            sort $negatives$ ascending by cross\_entropy($y_q$, $y_{pred}$ for each negative)\;
            take first $num\_positives$ and $num\_negatives$\;
        }\ElseIf{random sampling}{
            shuffle $positives$ and $negatives$\;
            take first $num\_positives$ and $num\_negatives$\;
        }

        \tcp{Compute loss component for this query}
        $loss_{query}$ = 0\;

        \For{$z_p \in positives$}{
            $loss_p$ = exp(sim($z_p$, $z_q$)/$\tau$)\;
            $loss_n$ = 0\;

            \For{$z_n \in negatives$}{
                $loss_n$ += exp(sim($z_n$, $z_q$)/$\tau$)\;
            }

            $loss_{query}$ += -log($loss_p$ / ($loss_p$ + $loss_n$))\;
        }

        $loss$ += $loss_{query}$ / len($positives$)\;
    }

    \textbf{return} $loss$ / len($queries$)\;
\end{footnotesize}
\caption{Contrastive Loss}
\label{alg:contrastive}
\end{algorithm}

\end{document}